\documentclass[10pt,twocolumn,letterpaper]{article}

\usepackage{adjustbox}
\usepackage{multirow}
\usepackage{booktabs}
\usepackage{iccv}
\usepackage{times}
\usepackage{epsfig}
\usepackage{graphicx}
\usepackage{caption}
\usepackage{subfigure}
\usepackage{makecell}

\usepackage{amsmath}
\usepackage{amssymb}

\usepackage{pifont}

\usepackage{amsfonts}
\usepackage{arydshln}
\usepackage{dsfont}

\usepackage{bbm}

\usepackage[normalem]{ulem}
\useunder{\uline}{\ul}{}

\newcommand{\cmark}{\ding{51}}%
\DeclareCaptionLabelSeparator{mycap}{.\hspace*{1.ex}}
\usepackage[pagebackref=true,breaklinks=true,letterpaper=true,colorlinks,bookmarks=false]{hyperref}

\iccvfinalcopy 

\def\iccvPaperID{9504} 

\ificcvfinal\fi

\begin{document}

\title{Online Class Incremental Learning on Stochastic Blurry Task Boundary \\ via Mask and Visual Prompt Tuning}

\author{
Jun-Yeong Moon\thanks{Equally contributed}$^*$\\
\and
Keon-Hee Park$^*$\\
\and
Jung Uk Kim$^\dagger$
\and
Gyeong-Moon Park\thanks{Corresponding author}\\
\and
Kyung Hee University, Yongin, Republic of Korea\\
{\tt\small \{moonjunyyy, pgh2874, ju.kim, gmpark\}@khu.ac.kr}
}

\maketitle

\begin{abstract}
   Continual learning aims to learn a model from a continuous stream of data, but it mainly assumes a fixed number of data and tasks with clear task boundaries. However, in real-world scenarios, the number of input data and tasks is constantly changing in a statistical way, not a static way. Although recently introduced incremental learning scenarios having blurry task boundaries somewhat address the above issues, they still do not fully reflect the statistical properties of real-world situations because of the fixed ratio of disjoint and blurry samples. In this paper, we propose a new Stochastic incremental Blurry task boundary scenario, called Si-Blurry, which reflects the stochastic properties of the real-world. We find that there are two major challenges in the Si-Blurry scenario: (1) intra- and inter-task forgettings and (2) class imbalance problem. To alleviate them, we introduce Mask and Visual Prompt tuning (MVP). In MVP, to address the intra- and inter-task forgetting issues, we propose a novel instance-wise logit masking and contrastive visual prompt tuning loss. Both of them help our model discern the classes to be learned in the current batch. It results in consolidating the previous knowledge. In addition, to alleviate the class imbalance problem, we introduce a new gradient similarity-based focal loss and adaptive feature scaling to ease overfitting to the major classes and underfitting to the minor classes. Extensive experiments show that our proposed MVP significantly outperforms the existing state-of-the-art methods in our challenging Si-Blurry scenario. The code is available at \url{https://github.com/moonjunyyy/Si-Blurry}
\vspace{-3mm}
\end{abstract}

\section{Introduction}

\begin{figure*}[h!]
    \centering
    \subfigure[Visualization of i-Blurry scenario \cite{koh2021online}.]{
        \includegraphics[width=\columnwidth]{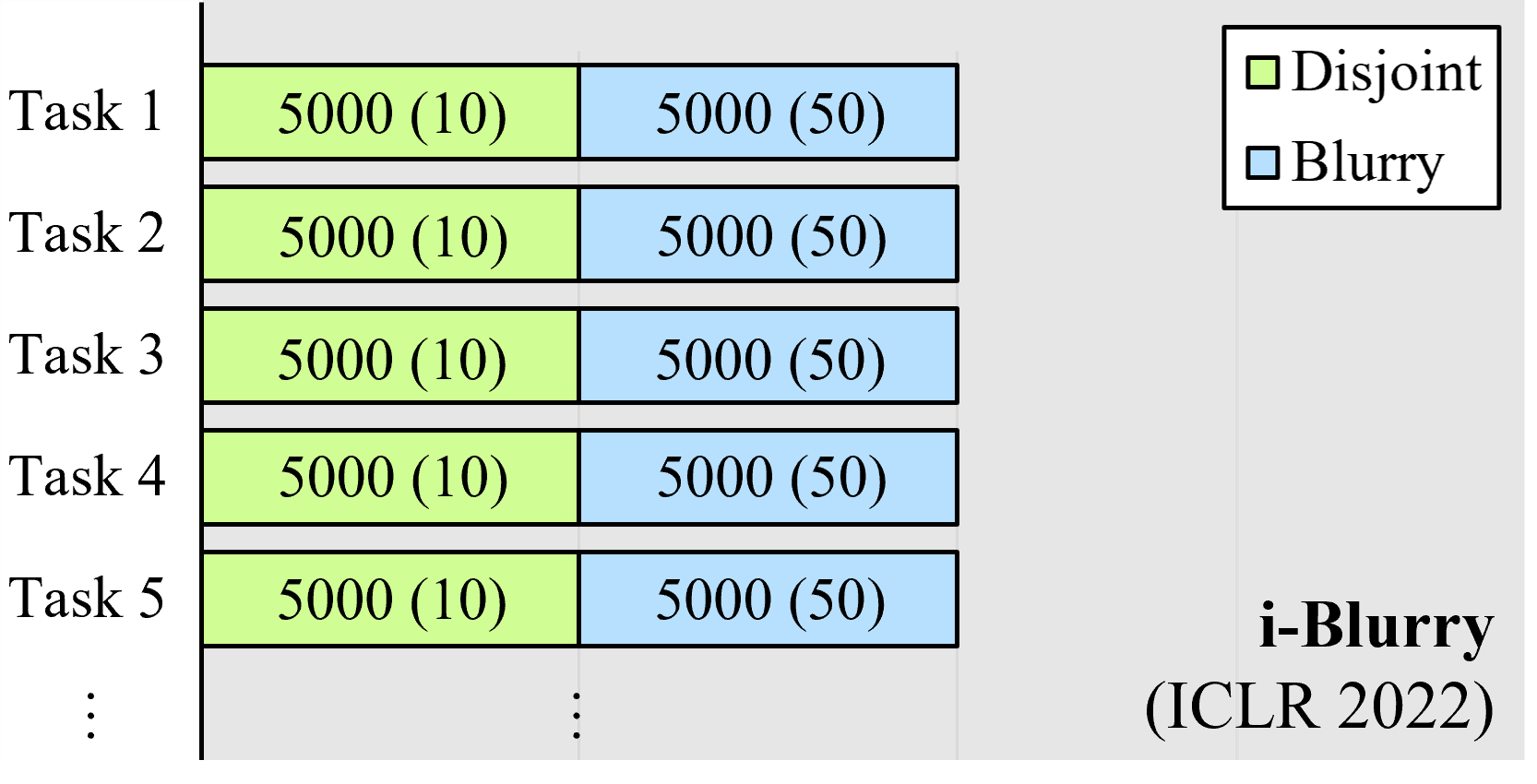}
        \label{fig:1-1}
    }
    \centering
    \subfigure[Visualization of our Si-Blurry scenario.]{
        \includegraphics[width=\columnwidth]{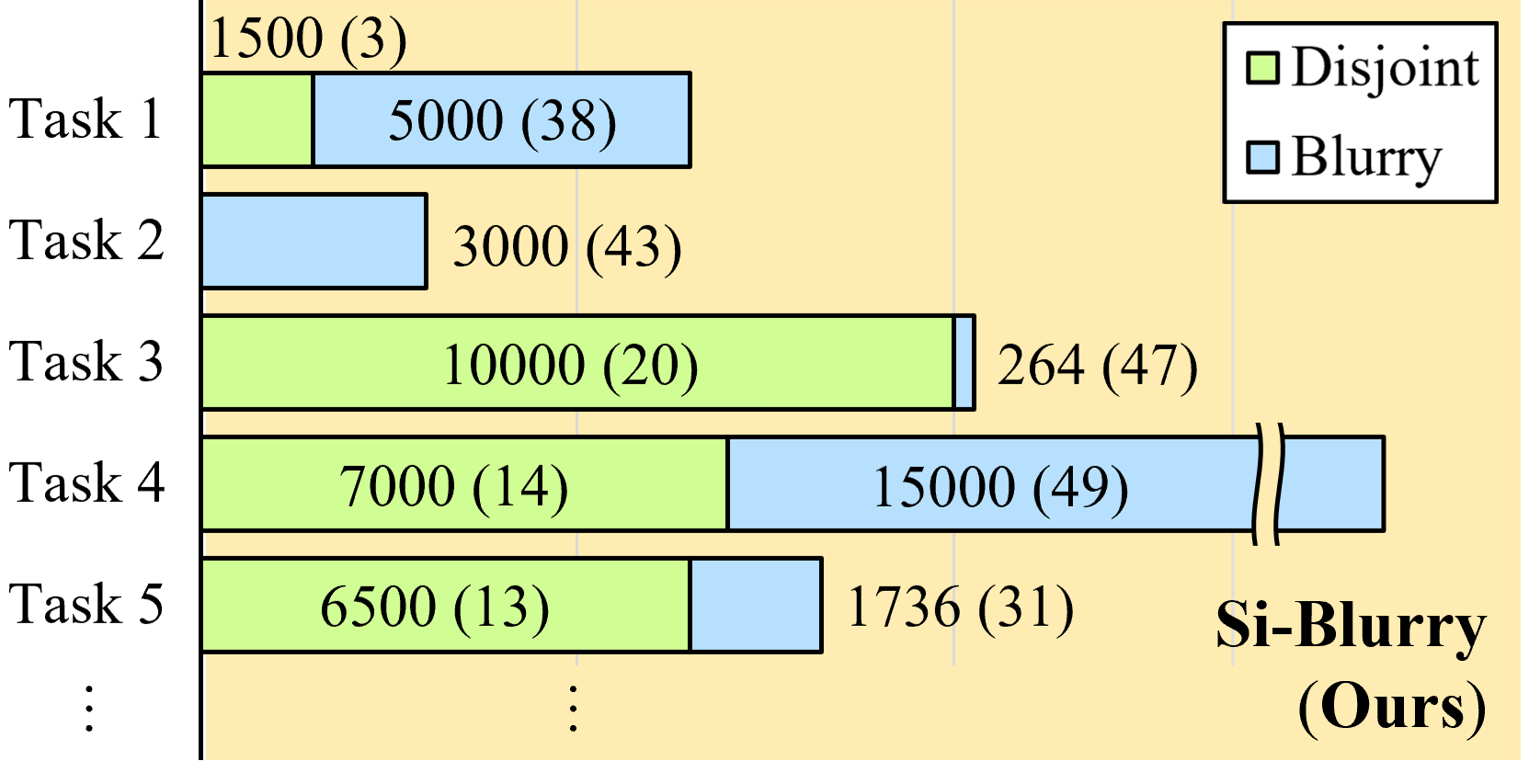}
        \label{fig:1-2}
    }
    \centering
    \subfigure[Nomalized search history from December 28, 2021, to December 28, 2022, NAVER Search Trend API : \url{https://www.ncloud.com/product/applicationService/searchTrend}.]{
        \includegraphics[width=\columnwidth]{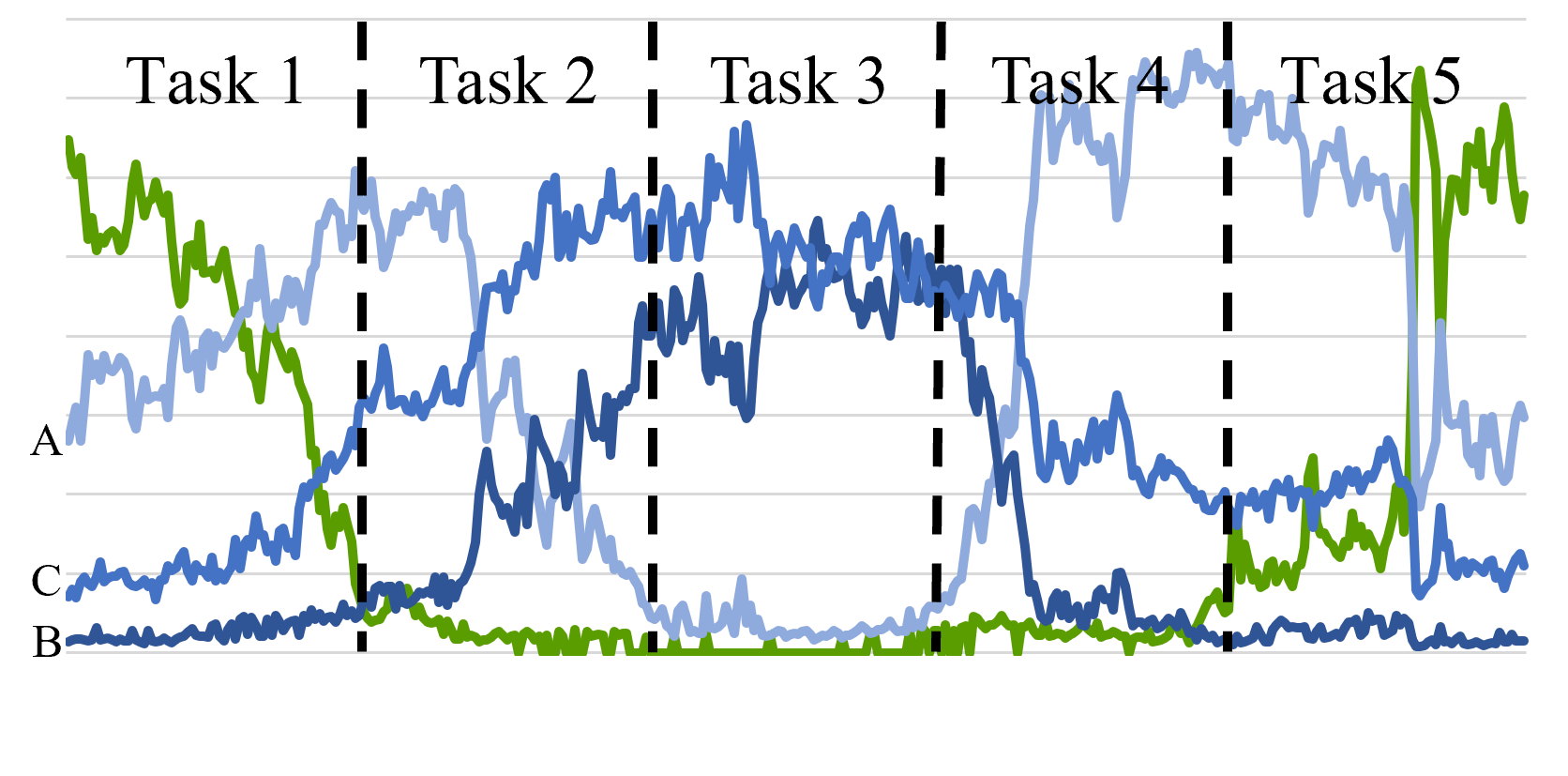}
        \label{fig:1-3}
    }
    \centering
    \subfigure[Example of task generated from real-world data.]{
        \includegraphics[width=\columnwidth]{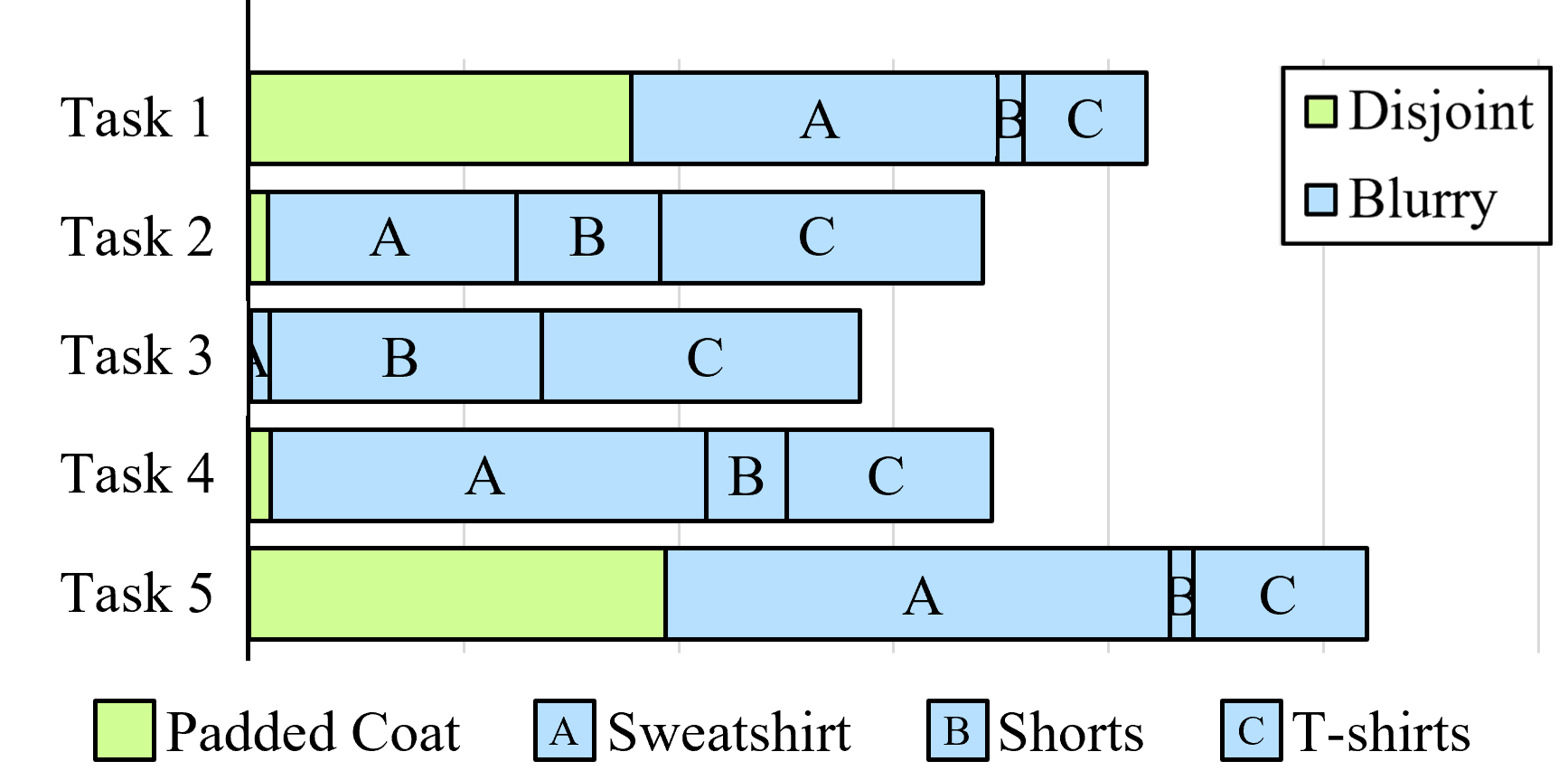}
        \label{fig:1-4}
    }
\vspace{-0.3cm}
\caption{Comparision of (a) i-Blurry scenario, (b) Si-blurry scenario, and (c), (d) real-world data. (a) i-Blurry scenario consists of the static number of classes in each task. all the tasks have the same number of disjoint classes and blurry classes. (b) The Si-Blurry scenario has an everchanging number of classes. The disjoint classes and blurry classes are different with each task. It denotes the unpredictable traits of the real-world.  (c) and (d) show the stream of real data.  Comparing (a), (b), and (d), Si-blurry is alike real-world task configuration.}
\vspace{-0.3cm}
\end{figure*}

Continual learning involves constantly learning from a stream of data while having limited access to previously seen information. In this scenario, unlike humans who can retain and apply their prior knowledge to new situations, modern deep neural networks face a challenge of \textit{catastrophic forgetting} \cite{MCCLOSKEY1989109, FRENCH1999128}. To overcome this challenge, various approaches are being explored \cite{pfulb2018a, parisi2019continual, gupta2020neural, lesort2020continual, de2021continual}. However, these traditional continual learning scenarios have clear task boundaries, where one can distinguish tasks with input data, unlike in the real-world. In real-world applications, clear task boundaries are often absent and access to data is limited to small portions at a time. This is referred to as online learning with blurry task boundaries \cite{bang2021rainbow}.

There are many cases of class emerging or disappearing like the stock market or e-commerce. To address this, the i-Blurry scenario \cite{koh2021online} has been recently proposed, which combines disjoint continual learning and blurry task-free continual learning. Although i-Blurry somewhat alleviates the above issue, it does not fully capture the complexity of real-world data, because i-Blurry has the fixed number of classes between tasks. Figure \ref{fig:1-1} shows the i-Blurry scenario that contains the static number of classes in each task. In the real-world scenarios, the number of classes and tasks vary dynamically as illustrated in Figure \ref{fig:1-3} and \ref{fig:1-4}. That is, as samples of a specific class continuously disappear or appear in the data stream, distribution of the data is dynamically changing.

To reflect the dynamic distribution of the real-world data, we propose a novel Stochastic incremental Blurry (Si-Blurry) scenario. We adopt a stochastic approach to imitate the chaotic nature of the real-world. As shown in Figure \ref{fig:1-2}, our Si-Blurry scenario is capable of effectively simulating not only newly emerging or disappearing data but also irregularly changing data distribution.
In the Si-Blurry scenario, we find that there are two main reasons for performance degradation: (1) intra- and inter-task forgettings, and (2) class imbalance problem. First, the continuous change of classes between batches causes intra- and inter-task forgettings, which make the model difficult to retain previously learned knowledge. Second, ignorance of minor classes and overfitting to major classes worsen the class imbalance problem in the Si-Blurry scenario. Minor class ignorance occurs by insufficient consideration of a few samples which belong to minor classes in training, and overfitting on major classes makes the model biased to a large number of samples that belong to major classes or disjoint classes.

\vspace{-1mm}
To deal with the aforementioned problems, we propose a novel online continual learning method called Mask and Visual Prompt tuning (MVP). We propose instance-wise logit masking and contrastive visual prompt tuning loss to alleviate intra- and inter-task forgettings by making classification easier and allowing prompts to learn the knowledge for each task effectively. Moreover, we propose a gradient similarity-based focal loss to prevent the problem of minor class ignorance. This method boosts learning of the ignored samples of minor classes in a batch, so that the samples for minor classes can be considered intensively. We also propose adaptive feature scaling to address the problem of overfitting to major classes. This method measures the marginal benefit \cite{cui2019class} of learning from a sample and prevents our model from learning already sufficiently trained samples.


We summarize our main contributions as follows:
\begin{itemize}
    \item We introduce a new incremental learning scenario, coined Si-Blurry, which aims to simulate a more realistic continual learning setting that the neural networks continually learn new classes online while a task boundary is stochastically varying.
    \item We propose an instance-wise logit masking and contrastive visual prompt tuning loss to prevent the model from intra-task and inter-task forgettings.
    \item To solve the class imbalance problem, we propose a new gradient similarity-based focal loss and adaptive feature scaling for minor-class ignorance and overfitting on major classes.
    \item We experimentally achieved significantly high performance compared to existing methods, supporting that our proposed method shows overwhelming performance and solves the problems of Si-Blurry in CIFAR-100, Tiny-ImageNet, and ImageNet-R.
\end{itemize}

\section{Related Work}
\subsection{Disjoint Continual Learning}
Class Incremental Learning (CIL) \cite{gepperth2016incremental} assumes that each task contains distinct classes not overlapping with another and that the class observed once in a task never appears again in subsequent tasks. CIL is categorized into the (1) regularization-based method, (2) replay-based method, (3) parameter isolation method, and (4) prompt-based method. The regularization-based method uses previous knowledge for regularizing the network while training new tasks \cite{li2017learning, kirkpatrick2017overcoming, castro2018end, liu2020more, wu2019large}. The replay-based method stores a few samples from the old task and replays them in the new task to mitigate catastrophic forgetting \cite{rebuffi2017icarl, wang2022foster, NEURIPS2019_e562cd9c, chaudhry2019tiny, lopez2017gradient}. The parameter isolation method expands the network or consists of sub-networks in a single network for each task \cite{shin2017continual, yan2021dynamically, aljundi2017expert, rusu2016progressive, yoon2017lifelong}. The prompt-based method proposed in natural language processing (NLP) for transfer learning attaches a set of learnable parameters, named prompt, to the frozen pre-trained model \cite{wang2022learning, wang2022dualprompt, jia2022visual}.

\subsection{Blurry Continual Learning}
Blurry Continual Learning \cite{prabhu2020gdumb,bang2021rainbow} assumes no new classes appear after the first task even though classes overlap across the tasks. A blurry setup has some requirements. First of all, each task is streamed sequentially. Second, the major class of each task is different. Last, a model can leverage only a small portion of data from the previous task. A blurry setup seems realistic. However, a blurry scenario has a shortage to apply real-world scenarios in that observing new classes is commonplace in a real-world scenario. i-Blurry \cite{koh2021online} proposes a more realistic setting that considers a blurry scenario with a class-incremental setting. However, the i-Blurry scenario also has a limitation of properly reflecting the real-world scenario due to: (1) the same number of new classes appearing in every task, and (2) new classes and blurry classes having the same proportion in every task. This is why we propose a stochastic incremental blurry scenario, which focuses on the stochastic property critical to real-world scenarios.

\subsection{Class Imbalance in Continual Learning}
Class imbalance, known as the long-tail problem, is that the classes are not represented equally in the classification task. Class imbalance is common in the real-world and it can cause inaccurate prediction performance in classification problems. In continual learning, the replay-based method suffers severe catastrophic forgetting due to the inequality between stored old samples and streamed new samples \cite{wu2019large}. To address this problem, existing methods consider gradient information to get the knowledge of prior tasks during training \cite{lopez2017gradient, NEURIPS2019_e562cd9c}, episodic memory management to enhance model performance by sampling effective samples \cite{chaudhry2018riemannian,liu2020mnemonics,welling2009herding}, and calibration of the bias \cite{wu2019large}. However, in a blurry scenario, incoming samples per class are different and this causes a class imbalance problem. Class imbalance in a task leads to bias for disjoint classes and major classes which exacerbates the training of the minority classes.

\section{Stochastic Incremental Blurry Scenario}

\begin{figure}[t!]
\begin{center}
     \centering
     \includegraphics[width=0.45\textwidth]{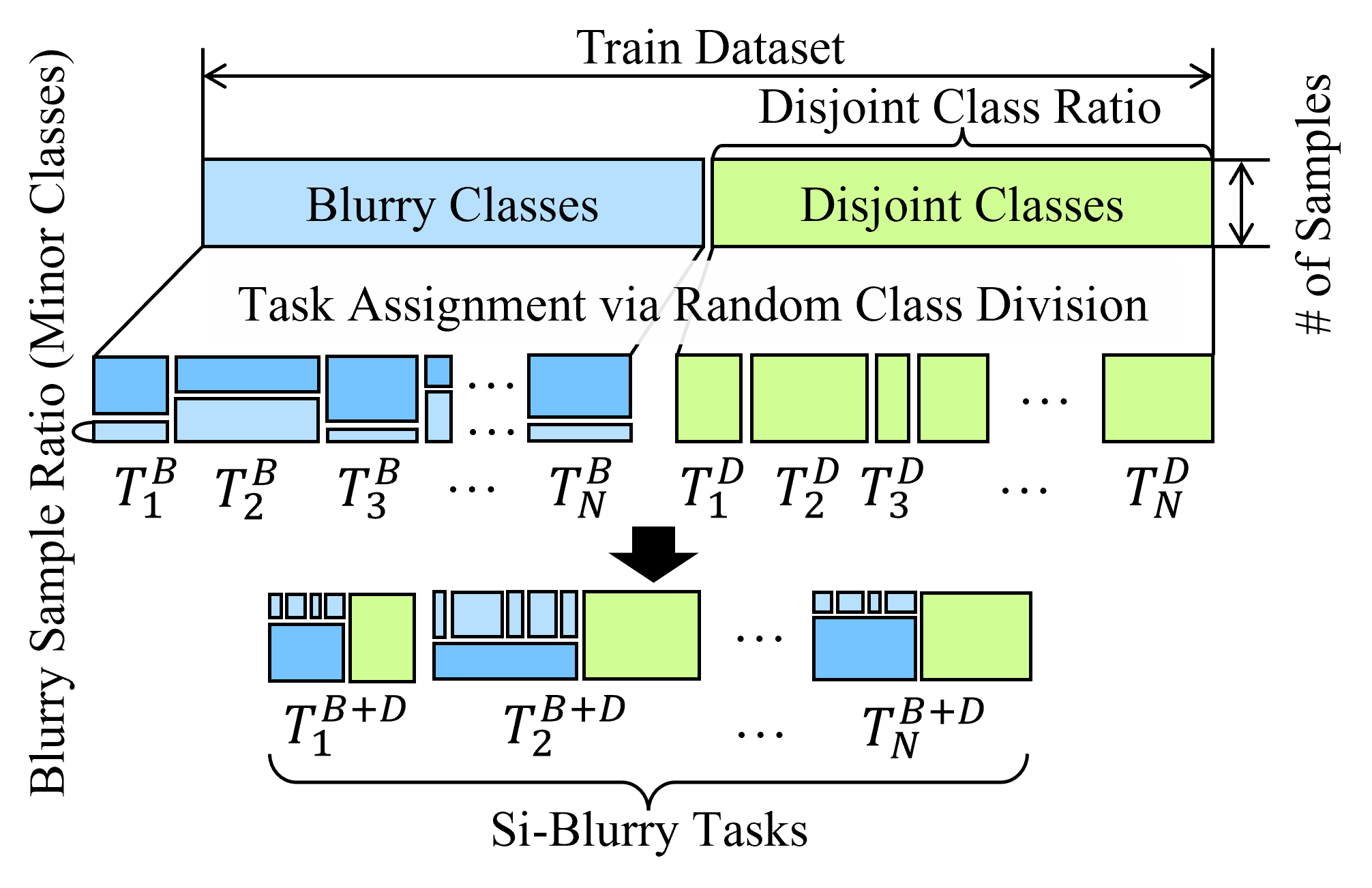}
     \vspace{-0.4cm}
     \caption{A configuration example of the proposed Si-Blurry scenario.}
\label{figure-2}
\vspace{-0.6cm}
\end{center}
\vspace{-0.3cm}
\end{figure}

\subsection{Scenario Configuration}
In a real-world scenario, the quantity of input data and tasks tends to change a stochastic manner. To simulate this, we propose the Stochastic incremental-Blurry (Si-Blurry) scenario. From \cite{koh2021online}, we divide the classes into two categories using disjoint class ratio: disjoint classes and blurry classes. As shown in Figure \ref{figure-2}, we randomly assign each blurry class and disjoint class to each blurry task ($T^B$) and disjoint task ($T^D$) by the disjoint class ratio. In blurry tasks, we gather the sample of blurry sample ratio and randomly distribute it to each task. This makes the classes on each task overlap, which blur explicit task boundary. Each task with a stochastic blurry task boundary $T^{B+D}$ consists of $T^B$ and $T^D$. Figure \ref{fig:1-2} shows an example of the distribution of Si-Blurry. Because the Si-Blurry task is stochastic, the batches get more diverse and imbalanced. As a result, there are lack of explicit task boundaries and the data imbalance, which pose significant challenges to formal continual learning methods. We define two problems that are exacerbated on Si-Blurry in the following subsections.

\begin{figure*}[h!]
\begin{center}
     \includegraphics[width=\textwidth]{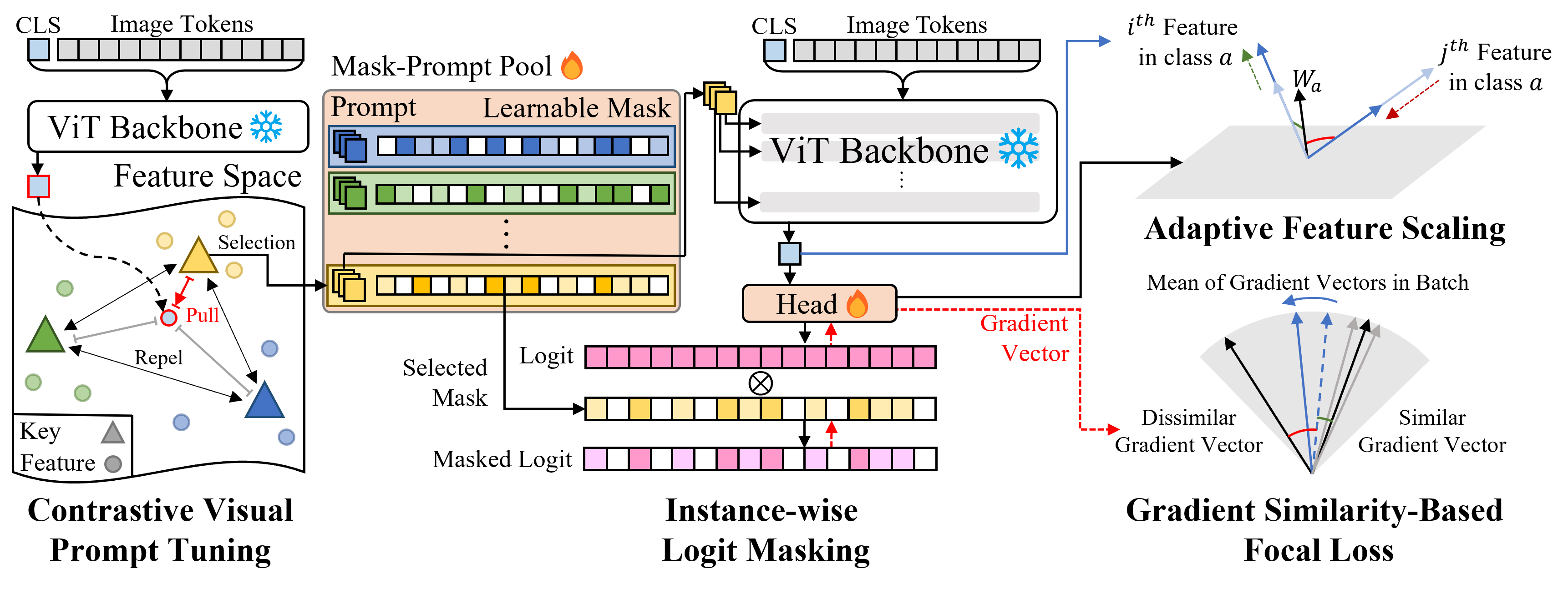}
\end{center}
    \vspace{-0.7cm}
    \caption{Illustration of the process of the proposed method. Firstly, a pre-trained ViT is used to extract features, and a key is compared with the feature by cosine distance to select the matched prompt and learnable mask. We use the prompt tuning \cite{wang2022dualprompt} method with deep prompting \cite{jia2022visual}. And the selected mask is multiplied element-wise to logit. To train the key, we use a contrastive manner and weighted key to prevent key grouping. After the forward pass, the weight vector of the label class in the classification head and feature are compared to measure the marginal benefit of the sample. In the backward path, we compare the mean of the gradient vector in the batch and the gradient vector of each sample to measure the ignored sample. Finally, by scaling the feature and the cross-entropy loss, we effectively handle the forgetting and class imbalance problems.}
    \label{fig:main}
    \vspace{-0.1cm}
\end{figure*}
\subsection{ Intra- and Inter-Task Forgettings }
\label{sub:3.2}

First, intra-task forgetting can be interpreted as inter-batch forgetting. This problem also presents in joint training but is largely addressed by randomizing the sample order in each batch. However, in online learning scenarios, each sample is only presented once, making it impossible to apply this strategy. Additionally, the stochastic nature of Si-Blurry creates more diverse batches, thereby intensifying the issue of intra-task forgetting. Intra-task forgetting can severely limit the ability of the model to learn and generalize well in the face of evolving and dynamic data distributions.

Inter-task forgetting, which refers to the phenomenon of losing previously learned knowledge of a task due to the changes in class distribution, is a major challenge in online continual learning. Si-Blurry, a scenario proposed to simulate the complexity of real-world data in a stochastic manner, does not have clear task boundaries like those in conventional continuous learning, making it difficult to handle this problem. The stochastic change in class distribution in Si-Blurry exacerbates this problem. Effective strategies need to be devised to enable the model to learn the knowledge continually from data with varying class distributions, without losing previously acquired knowledge, and any catastrophic forgetting.

\subsection{Class Imbalance}
\label{sub:3.3}

Minor-class ignorance and overfitting to major classes cause the class imbalance problem. The issue of minor-class ignorance arises when the number of samples in a batch varies and this issue causes an imbalanced weighted loss. The loss imbalance leads to the negligence of the minority of samples in a batch, resulting in their poor representation while training the model. Overfitting to major classes is, conversely, a phenomenon where a class with a large number of samples deteriorates the model performance by acquiring unnecessary knowledge relating to generalization performance.

These are particularly important issues in the context of Si-Blurry, where no explicit task boundaries exist, and continual learning requires a model to capture the knowledge from a wide range of samples. Finding a solution to this problem is essential for ensuring that the model remains effective in recognizing and classifying all samples, regardless of the number of samples per class, and that it continues to learn and improve over time.

\section{Mask and Visual Prompt Tuning (MVP)}

\subsection{Preliminary and Problem Formulation}
The Si-Blurry scenario considers learning a model with only a few samples because it cannot access the whole current training data. When given the accessible data $\mathcal{B}=\left\{\mathbf{x_i},y_i \right\}_{i=1}^N$, where $\mathbf{x_i}\in \mathcal{X}$, $y_i \in \mathcal{Y}$, we reshape the samples to a flattened patch shape $\mathbb{R}^{L \times (S^2 \times C)}$ to feed into the pre-trained model $ f: \mathbb{R}^{L \times (\mathrm{S}^2 \times \mathrm{C})} \to \mathbb{R}^{L \times \mathrm{D}} $ which is frozen, where $L$, $S$, $C$, and $D$ represent the token length, patch size, channel, and embedding dimension, respectively. A linear classifier $W\in \mathbb{R}^{D \times \left| \mathcal{Y} \right|}$ is trainable.

Previous studies demonstrate the effectiveness of using knowledge from the pre-trained model and tuning the small size of parameters \cite{wang2022learning, wang2022dualprompt} for continual learning. To this end, we adopt the prompt tuning method \cite{lester2021power} for online continual learning. Similarly to DualPrompt \cite{wang2022dualprompt}, we utilize a pre-trained Vision Transformer (ViT) as a feature extractor for the query. We match the query with the key to apply the contrastive visual prompt tuning loss and select the prompt.

\subsection{Instance-wise Logit Masking}

Existing prompt methods assume the explicit task boundary and require information on the task boundary for training, which is not feasible in Si-Blurry, where no explicit task boundary exists. The cross-entropy loss is highly effective in optimizing classification models, but it requires a proper comparison target to acquire sufficient knowledge. Unlike traditional joint training, the absence of comparison constantly leads to forgetting in online continual learning. To address this problem causing intra- and inter-task forgettings, we propose a new instance-wise logit masking technique.

To complement the prompt-based continual learning approach and further enhance the performance of the model, we introduce a learnable mask paired with prompts that helps the model to learn more intra-relevant and easier learning goals. Since the key-value mechanism is used to select each prompt, where a feature extracted by the pre-trained model serves as a query, each prompt can be responsible for a certain region of the feature space in which classes have similar extracted features. As illustrated in Figure \ref{fig:main}, we apply the mask to logit using an element-wise product and train the mask, and then calculate cross-entropy loss which makes the mask divide the tasks into easier classification tasks. The logit masking provides the model with a scaled gradient during back-propagation, which protects the knowledge that has been sufficiently trained and encourages the learning of classes to be learned.

\subsection{Contrastive Visual Prompt Tuning Loss}

The logit mask assumes that each key enables the prompts to learn similar knowledge. We empirically find that the existing prompt-based method \cite{wang2022learning} converges to a single point, which renders the query selection mechanism inaccurate and meaningless. Moreover, the keys are updated continuously, which causes forgetting. To overcome these challenges and leverage the benefits of prompt-based continual learning, we propose a novel loss function, called Contrastive Visual Prompt Tuning Loss. We formulate this loss term as follows:
\begin{align}
    \label{Eq-4}
    & s_p = \sum^{P}_{p=1}\sum^{B}_{q=1}{\mathrm{exp}\left( \delta \left( \textbf{k}_p,\textbf{q}_q \right)/(\mathcal{C}_p +1)\right)}, \nonumber\\
    & s_n = \sum^{P}_{p=1}\sum^{P}_{q=1}{\mathrm{exp}\left( \delta \left( \textbf{k}_p,\textbf{k}_q \right)/(\mathcal{C}_p +1)\right)}, \nonumber\\
    &\mathcal{L}_{CVPT} \, = \, -\mathrm{log}\, \frac{ s_n}{ s_p+ s_n},
\end{align}
where $\delta$ is cosine distance, $P$ denotes the size of the prompt pool, $\textbf{q}_n\in\mathbb{R}^{D}$ indicates the query feature, $\textbf{k}_n\in\mathbb{R}^{D}$ denotes the key of $n^{th}$ prompt, and $\mathcal{C}_n$ means the count of selection of $n^{th}$ prompt. In $\mathcal{L}_{CVPT}$, $(\mathcal{C}_p +1)$ plays a role of the temperature to control the softness. If $\mathcal{C}_n$ is large, the effect of loss to key lessens. As illustrated in Figure \ref{fig:main}, the $\mathcal{L}_{CVPT}$ increases the distances between keys. Also, as the prompt learns, the keys become heavier to ensure consistency in key selection. The instance-wise logit masking coupled with $\mathcal{L}_{CVPT}$ can prevent inter-task and intra-task forgetting by ensuring that each prompt divides its responsible region and preserves the knowledge.



\subsection{Gradient Similarity-based Focal Loss}
Since the blurry setup has a task that comprises of imbalanced classes, it is challenging for the model to extract the knowledge of all the observed classes in the blurry setup.
Due to the stochastic nature of Si-Blurry, we cannot guarantee a minimum number of samples for minor classes. 
To mitigate the aforementioned minor class ignorance, we propose a Gradient Similarity-based Focal loss $\mathcal{L}_{GSF}$ (GSF loss). It focuses on the loss from ignored samples leveraging ignore scores $\mathrm{Score}^{ign}$. The ignore score $\mathrm{Score}^{ign}_i$ denotes how much a sample $\mathbf{x_i}$ is ignored by other samples during training. We use cosine distance in between a gradient vector from each sample  $\nabla W_{y_i}(f(\mathbf{x_i}))\in\mathbb{R}^{D}$ and the averaged gradient vector $\nabla W_{y_i}(f(\mathcal{B}))\in\mathbb{R}^{D}$ from accessible data $\mathcal{B}$ to yield an ignore score for a sample $\textbf{x}_i$. We formulate ignore score and GSF loss as:
\begin{align}
    \label{Eq-5}
    &\nabla W_{y_i}(f(\mathcal{B}))=\frac{1}{\left|\mathcal{B} \right|}\sum_{(\textbf{x},y) \in \mathcal{B}}\nabla W_{y_i}(f(\textbf{x})), \nonumber \\
    &\mathrm{Score}_{i}^{ign} = \delta\left( \nabla{W_{y_i} \left(f\left(\textbf{x}_i\right)\right)},\,\nabla W_{y_i}(f(\mathcal{B})\right)), \\
    \label{Eq-6}
    &\mathcal{L}_{GSF} =\frac{1}{\left|\mathcal{B} \right|}\sum_{i=1}^{\left|\mathcal{B} \right|}\left ( \mathrm{Score}_{i}^{ign} \right )^\gamma\cdot  \mathcal{L}_{CE}\left ( \hat{y_{i}},y_{i} \right ),
\end{align}
where $(\textbf{x},y)\in \mathcal{B}$ is a training sample, $\mathcal{L}_{CE}$ is a cross-entropy loss, and $W_{y_i}$ is the weights of corresponding label, respectively. Eq. \ref{Eq-5} represents ignore score for a sample $\textbf{x}_i$. When the $\mathrm{Score}^{ign}$ has a high value, it implies the model is hard to extract the knowledge from the sample, whereas the low $\mathrm{Score}^{ign}$ means vice versa. Leveraging $\mathrm{Score}^{ign}$, we can emphasize the loss from the ignored sample and capture more knowledge of minor classes than before as illustrated in Figure \ref{fig:main}. Eq. \ref{Eq-6} represents GSF loss which considers the amount of ignorance. In \cite{lin2017focal}, the focal loss dynamically scales cross-entropy loss considering confidence in the correct class. Our proposed GSF loss also dynamically scales cross-entropy loss. However, our loss scales the cross-entropy loss considering ignore score. Ignore score is the degree of ignorance that is conceptually different from confidence. GSF loss can mitigate the class ignorance problem which minor classes highly suffered, and enables balanced class learning.


\subsection{Adaptive Feature Scaling}
In online learning, the model cannot access all the data of the current task but access a few samples. In our novel Si-Blurry scenario, each task has a class imbalance problem. When the model access training data, there are no or few samples of minor classes. It makes the model overfit to major classes and newly streamed disjoint classes. To mitigate the overfitting problem caused by the class imbalance, we propose Adaptive Feature Scaling (AFS) which expands or contracts the feature vector considering the marginal benefit score $\mathrm{Score}^{MB}$. Using the $\mathrm{Score}^{MB}$, the model can learn new knowledge from the accessible data while preserving the knowledge from inaccessible prior data.

As prior works \cite{deng2019arcface,liu2017sphereface,liu2016large} suggest, the similarity between the feature vector and the weights from the last fully connected layer relates to the prediction when the model is trained by cross-entropy loss with softmax function. 
We propose a marginal benefit score $\mathrm{Score}^{MB}$ that represents how similar the feature vector is with the weights of the corresponding label. Leveraging this, we estimate the marginal benefit from the given instance and adjust the model updates by the given instance. We can calculate $\mathrm{Score}^{MB}$ as follows:
\begin{align}
    &\mathrm{Score}^{MB}_i = \delta\left( f\left( \textbf{x}_i\right),W_{y_i} \right)\ + \textit{m}, \\
    &\mathbf{h_i} = \frac{f({\textbf{x}_i})}{\mathrm{Score}^{MB}_i}, \\
    &\widehat{y}_i = W(\mathbf{h_i})
\end{align}
where $\textit{m}$ is a margin, and $\mathbf{h_i}$ is a scaled feature vector by $\mathrm{Score}^{MB}_i$. When the $\mathrm{Score}^{MB}$ has a high value, it implies the given sample has a large marginal benefit. The $\mathrm{Score}^{MB}$ reduces the feature vector to increase the expected loss value. Enlarged expected loss makes the model learn enough knowledge from the given sample. In contrast, when the $\mathrm{Score}^{MB}$ has a low value, it implies the given sample has a little marginal benefit for the model. In this case, a feature vector is expanded to decrease the expected loss value. The model is less trained due to the curtailed expected loss. We estimate the marginal benefit that can be extracted from a sample and scale up and down the feature vector considering the marginal benefit.

To overcome the class imbalance problem, we propose two components that seem similar: gradient similarity-based focal loss (GSF) and adaptive feature scaling (AFS). Although GSF and AFS look similar, their main roles are different. GSF emphasizes learning minor classes to tackle the class imbalance problem in the task. AFS regularizes learning major classes to address the overfitting problem.

Finally, we train our model in an end-to-end manner. The total loss for our method is defined as:
\begin{align}
    &\mathcal{L}_\mathrm{total} = (1-\alpha)\mathcal{L}_\mathrm{CE} + \alpha\mathcal{L}_{GSF} + \mathcal{L}_{CVPT}\,,
\end{align}
where $\mathcal{L}_\mathrm{CE}$ is cross entropy loss with instance-wise logit masking. We use hyperparameter $\alpha$ for the balanced training and $\gamma$ at the gradient similarity-based focal loss to scale the ignore score.


\begin{table*}[h!]
\begin{center}
\renewcommand{\tabcolsep}{1.8mm}
\resizebox{\textwidth}{!}{
\begin{tabular}{ccccccccccc}
\specialrule{1.1pt}{1pt}{1pt}
\multirow{2}{*}{\begin{tabular}[c]{@{}c@{}}\textbf{Buffer} \\ \textbf{Size}\end{tabular}} & \multirow{2}{*}{\textbf{Method}} &  & \multicolumn{2}{c}{\textbf{CIFAR-100}} &  & \multicolumn{2}{c}{\textbf{Tiny-ImageNet}} &  & \multicolumn{2}{c}{\textbf{ImageNet-R}}  \\ \cline{4-5} \cline{7-8} \cline{10-11}
 &  &  & $A_\mathrm{AUC}$ & $A_\mathrm{Last}$ &  & $A_\mathrm{AUC}$ & $A_\mathrm{Last}$ &  & $A_\mathrm{AUC}$ & $A_\mathrm{Last}$  \\ \hline
 
\multicolumn{1}{c|}{\multirow{6}{*}{\textbf{0}}} & Finetuning &  & 19.71±3.39 & 10.42±4.92 &  & 15.50±0.74 &  10.42±4.92 &  & 7.51±3.94 & 2.29±0.85  \\
\multicolumn{1}{c|}{} & Linear Probing &  & 49.69±6.09 & 23.07±7.33 &  & 42.15±2.79 & 21.97±6.43 &  & 29.24±1.26 & 16.87±3.14  \\
\multicolumn{1}{c|}{} & LwF \cite{li2017learning} &  & 55.51±3.49 & 36.53±10.96 &  & 49.00±1.52 & 27.47±7.59 &  & 31.61±1.53 & 20.62±3.67  \\
\multicolumn{1}{c|}{} & L2P \cite{wang2022learning} &  & 57.08±4.43 & 41.63±12.73 &  & 52.09±1.92 & 35.05±5.73 &  & 29.65±1.63 
 & 19.55±4.78  \\
\multicolumn{1}{c|}{} & DualPrompt \cite{wang2022dualprompt} &  & \underline{67.07±4.16} & \underline{56.82±3.49} &  & \underline{66.09±2.00} & \underline{48.72±3.41} &  & \underline{40.11±1.27} & \underline{29.24±4.63}  \\ \cdashline{2-11} 
 \multicolumn{1}{c|}{}\rule{0pt}{10pt}  & \textbf{MVP (Ours)} &  &  \textbf{68.10±4.91}  &  \textbf{62.59±2.38}  &  &  \textbf{68.95±1.33}  &  \textbf{52.78±2.08} &  & \textbf{40.60±1.21} & \textbf{31.96±3.07} 
\\ \hline

\multicolumn{1}{c|}{\multirow{5}{*}{\textbf{500}}} & ER \cite{rolnick2019experience} &  & 65.57±4.77 & 60.68±1.15 &  & 59.46±1.81 & 40.60±2.71 &  & \underline{40.31±1.33} & 28.85±1.43 \\
\multicolumn{1}{c|}{} & EWC++ \cite{kirkpatrick2017overcoming} &  & 34.54±5.19 & 25.62±3.35 &  & 55.05±1.75 & 34.88±3.65   & & 18.62±1.00 & 11.36±2.40 \\
\multicolumn{1}{c|}{} & RM \cite{bang2021rainbow} &  & 40.86±3.32 & 23.94±0.61 &  & 31.96±0.80 & 7.43±0.27 &  & 18.31±1.09 & 4.14±0.18  \\
\multicolumn{1}{c|}{} & CLIB \cite{koh2021online}&  & \underline{69.68±2.20} & \underline{67.16±0.72} &  & \underline{60.11±1.53} & \underline{48.97±1.48} &  & 37.18±1.52 & \underline{29.51±0.98} \\ \cdashline{2-11}
 \multicolumn{1}{c|}{}\rule{0pt}{10pt} & \textbf{MVP-R (Ours)} &  &  \textbf{76.06±4.22}  &  \textbf{79.32±1.28}  &  &  \textbf{76.52±0.73}  &  \textbf{65.19±0.58} & & \textbf{49.07±1.47} & \textbf{44.17±1.72} \\ 
\hline

\multicolumn{1}{c|}{\multirow{5}{*}{\textbf{2,000}}} & ER \cite{rolnick2019experience} &  & 69.86±4.08 & 71.81±0.69 &  & \underline{66.75±1.13} & 55.07±1.28 &  & \underline{45.74±1.35} & \underline{38.13±0.32}  \\
\multicolumn{1}{c|}{} & EWC++ \cite{kirkpatrick2017overcoming} &  & 47.75±5.35 & 46.93±1.44 &  & 64.92±1.21 & 53.04±1.53 &  & 30.20±1.31 & 21.28±1.88 \\
\multicolumn{1}{c|}{} & RM \cite{bang2021rainbow} &  & 53.27±3.00 & 65.51±0.55 &  & 47.26±1.13 & 44.55±0.37&  & 27.88±1.29 & 24.25±0.99   \\
\multicolumn{1}{c|}{} & CLIB \cite{koh2021online} &  & \underline{71.53±2.61} & \underline{72.09±0.49} &  & 65.47±0.76 & \underline{56.87±0.54} &  & 42.69±1.30 & 35.43±0.38  \\ \cdashline{2-11} 
 \multicolumn{1}{c|}{}\rule{0pt}{10pt} & \textbf{MVP-R (Ours)} &  & \textbf{78.65±3.59} & \textbf{84.42±0.44} &  & \textbf{80.67±0.75} & \textbf{74.34±0.32} &  &  \textbf{52.47±1.45} &\textbf{50.54±2.08}  \\ 
\specialrule{1.1pt}{1pt}{1pt}
\end{tabular}
}
\end{center}
\vspace{-0.6cm}
    \caption{Average accuracy of continual learning methods on Si-Blurry scenario. For the comparison, we adopt the CIFAR-100, Tiny-ImageNet, and ImageNet-R datasets. Note that MVP-R indicates our MVP with a memory buffer.}
\label{table-main}
\end{table*}

\begin{table*}[h!]
\begin{center}
\begin{tabular}{c|cc|cc|cc}
\specialrule{1.1pt}{1pt}{1pt}
\multirow{2}{*}{\textbf{Method}} & \multicolumn{2}{c|}{\textbf{Components}} & \multicolumn{2}{c|}{\textbf{Memory = 0}}    & \multicolumn{2}{c}{\textbf{Memory = 2,000}} \\ \cline{2-7}
                        & \multicolumn{1}{c}{Mask} & Cont & $A_\mathrm{AUC}$ & $A_\mathrm{Last}$ & $A_\mathrm{AUC}$ & $A_\mathrm{Last}$           \\ \hline
Baseline                & \textbf{-} & \textbf{-} & 67.07±4.16 & 56.82±3.49 & 75.26±5.02 & 80.72±0.83 \\ \hline
\multirow{3}{*}{\textbf{MVP (Ours)}} & \cmark &   & 67.64±3.81 & 58.52±3.28 & 77.83±0.35 & 84.26±0.04 \\
                        &  & \cmark         & 66.37±4.59 & 58.63±1.18 & 76.67±1.98 & 83.32±0.40 \\
                        & \cmark & \cmark   & \textbf{68.08±6.46} & \textbf{60.20±3.28} & \textbf{77.85±0.04} & \textbf{84.28±0.15} \\
\specialrule{1.1pt}{1pt}{1pt}
\end{tabular}%
\end{center}
\begin{center}

\begin{tabular}{c|cc|cc|cc}
\specialrule{1.1pt}{1pt}{1pt}
\multirow{2}{*}{\textbf{Method}} & \multicolumn{2}{c|}{\textbf{Components}} & \multicolumn{2}{c|}{\textbf{Memory = 0}}    & \multicolumn{2}{c}{\textbf{Memory = 2,000}} \\ \cline{2-7}
                        & \multicolumn{1}{c}{GSF} & AFS & $A_\mathrm{AUC}$ & $A_\mathrm{Last}$ & $A_\mathrm{AUC}$ & $A_\mathrm{Last}$ \\ \hline
Baseline                & \textbf{-} & \textbf{-} & 67.07±4.16 & 56.82±3.49 & 75.26±5.02 & 80.72±0.83 \\ \hline
\multirow{3}{*}{\textbf{MVP (Ours)}} & \cmark &      & 67.45±5.21 & 56.11±3.20 & 77.34±2.16 & 83.75±0.53 \\
                        &  & \cmark         & 67.45±3.78 & 57.93±2.11 & 77.86±2.09 & 84.31±0.20 \\
                        & \cmark & \cmark   & \textbf{67.66±3.47} & \textbf{ 58.28±2.95} & \textbf{78.28±3.67} & \textbf{84.41±0.21} \\
\specialrule{1.1pt}{1pt}{1pt}
\end{tabular}%
\end{center}
\vspace{-0.4cm}
\caption{Ablation studies of instance-wise logit masking and the contrastive visual prompt tuning for alleviating inter and intra-task forgetting. `Mask' and `Cont' represent instance-wise logit masking and contrastive visual prompt tuning, respectively. The lower table represents the ablation study of gradient similarity-based focal loss and adaptive feature scaling. GSF and AFS denote gradient similarity-based focal loss and adaptive feature scaling, respectively. }
\label{table-ablation}
\end{table*}

\begin{table*}[h!]
\begin{center}
\begin{tabular}{c|cccc|cc|cc}
\specialrule{1.1pt}{1pt}{1pt}
\multicolumn{1}{c|}{\multirow{2}{*}{\textbf{Method}}}   & \multicolumn{4}{c|}{\textbf{Components}}    & \multicolumn{2}{c|}{\textbf{Memory = 0}} & \multicolumn{2}{c}{\textbf{Memory = 2,000}} \\ \cline{2-9} 
                            & \textbf{Mask} & \textbf{Cont} & \textbf{GSF} & \textbf{AFS} & $A_\mathrm{AUC}$ & $A_\mathrm{Last}$ & $A_\mathrm{AUC}$ & $A_\mathrm{Last}$ \\ \hline
Baseline                    & \textbf{-} & \textbf{-} & \textbf{-} & \textbf{-} & 67.07±4.16 & 56.82±3.49 & 75.26±5.02 & 80.72±0.53 \\ \hline
\multirow{3}{*}{\textbf{MVP (Ours)}} & \cmark & \cmark &     &           & 68.08±6.46 & 60.20±3.28 & 77.85±0.04 & 84.28±0.15 \\
                            &  &  & \cmark  & \cmark            & 67.66±3.47 & 58.28±2.95 & 78.28±3.67 & 84.41±0.21 \\
                            & \cmark & \cmark & \cmark & \cmark & \textbf{68.10±4.91} & \textbf{62.59±2.38} & \textbf{78.65±3.59} & \textbf{84.42±0.44} \\
\specialrule{1.1pt}{1pt}{1pt}
\end{tabular}
\end{center}
\vspace{-0.4cm}
\caption{Ablation experiment coupled with the best condition for inter and intra-task forgetting and class imbalance. `Mask' and `Cont' represent instance-wise logit masking and contrastive visual prompt tuning and GSF and AFS denote gradient similarity-based focal loss and adaptive feature scaling, respectively.}
\label{table-best_ablation}
\end{table*}

\section{Experiments}
\subsection{Experimental Details}
\noindent \textbf{Datasets.} We conducted experiments on three datasets: CIFAR-100 \cite{krizhevsky2009learning}, Tiny-ImageNet \cite{le2015tiny}, and Imagenet-R \cite{hendrycks2021many} with 60,000, 100,000, and 30,000 samples, respectively, and 100, 200, and 200 classes. For the Si-Blurry scenario, we set a disjoint class ratio to 50\% and a blurry sample ratio to 10\%. We evaluated each method on five independent seeds to empirically search the difficulty of Si-Blurry. \\

\noindent \textbf{Baselines.} We compared our method with the state-of-the-art methods, including replay-based methods such as ER \cite{rolnick2019experience}, Rainbow Memory (RM) \cite{bang2021rainbow}, and CLIB \cite{koh2021online}, regularization-based method LwF \cite{li2017learning}, replay-regularization combined method EWC++ \cite{kirkpatrick2017overcoming}, and prompt-based methods L2P \cite{wang2022learning} and DualPrompt \cite{wang2022dualprompt}. 
We used finetuning and linear probing as lower-bound of our scenario. We implemented all methods with the fixed pre-trained Vision Transformer \cite{dosovitskiy2020image}, but the EWC++ and finetuning conducted full finetuning. \\

 \noindent \textbf{Implementation Details.} We used Adam optimizer \cite{adam} to train our MVP with 0.005 learning rate. For the existing methods \cite{rolnick2019experience,bang2021rainbow,koh2021online,li2017learning,kirkpatrick2017overcoming,wang2022learning,wang2022dualprompt}, we re-implemented them by following the setting (\textit{e.g.,} learning rate) of original papers in order to compare the performance with our MVP in the Si-Blurry scenario. We set the batch size to 32. For training our MVP, we set $\alpha$ to $0.5$ and $\gamma$ to $2.0$. We empirically observed that our method is robust to $\alpha$ and $\gamma$.
 Please note that although our MVP is designed to be memory-free, it also can use a memory buffer if it is needed. To this end, we adopt the replay buffers capable of storing 500 and 2,000 samples. We call our MVP with replay buffers ``MVP-R''.\\

 
\noindent  \textbf{Evaluation Metrics.} To evaluate the online learning performance, we used two metrics, $A_\mathrm{AUC}$ and $A_\mathrm{Last}$. $A_\mathrm{AUC}$ metric, proposed in \cite{koh2021online}, measures the performance of anytime inference. Anytime inference assumes that inference queries can occur anytime during training with exposed classes. $A_\mathrm{Last}$ measures the inference performance after the train is ended. In the real-world, the model needs to offer the proper services whenever a client is needed. In this point of view, Evaluating performance by $A_\mathrm{Last}$ and $A_\mathrm{AUC}$ is appropriate to benchmark the performance in the real-world.

\subsection{Results on the Si-Blurry Scenario}
We compared our MVP to other online CL methods in the Si-Blurry scenario. The results are shown in Table \ref{table-main}. In the Si-Blurry scenario, our MVP method outperformed all other comparison methods. On CIFAR-100, our method without a memory buffer outperformed EWC++ and RM, where they used a memory buffer size of 500. Furthermore, our method with a memory buffer size of 500 overwhelmed other methods with large margins. On Tiny-ImageNet, our MVP also outperformed other methods. Note that the RM Score was especially low on Tiny-ImageNet. We believe that the RM method trains the model highly focusing on samples stored in a memory buffer and Tiny-ImageNet contains 100,000 training images. The limited memory buffer could not cover the vast number of samples. So RM method suffered from overfitting. On ImageNet-R, all comparison methods scored low performance whether a memory buffer was used or not. Nonetheless, MVP showed a standing-out performance score. The score gap between MVP and other methods got bigger when the memory buffer was used. Our method MVP got an outperforming score without memory and performed much better when memory is used.

\begin{table*}[h!]
\begin{center}
\begin{tabular}{ccccccc}
\specialrule{1pt}{1pt}{1pt}
\multirow{2}{*}{\textbf{Case}} & \multicolumn{3}{c|}{\textbf{i-Blurry}} & \multicolumn{3}{c}{\textbf{Si-Blurry}} \\ \cline{2-4} \cline{5-7} 
            & \multicolumn{1}{c}{CLIB \cite{koh2021online}} & \multicolumn{1}{c}{DP\cite{wang2022dualprompt}} & \multicolumn{1}{c|}{MVP-R (2,000)} &  \multicolumn{1}{c}{CLIB \cite{koh2021online}} & \multicolumn{1}{c}{DP \cite{wang2022dualprompt}} & MVP-R (2,000) \\ \hline
Best case    & 72.56 & 67.51 & \multicolumn{1}{c|}{84.69} & 72.91 & 61.68 & 84.89 \\
Worst case   & 71.86 & 62.57 & \multicolumn{1}{c|}{83.68} & 71.78 & 53.68 & 83.80\\ \cdashline{1-7}
Average      & 72.12±0.38 & 64.90±1.96 & \multicolumn{1}{c|}{\textbf{84.44±0.43}} & 72.09±0.49 & 56.82±3.49 & \textbf{84.42±0.44} \\
\specialrule{1pt}{1pt}{1pt}
\end{tabular}
\end{center}
    \vspace{-5mm}
    \caption{The analysis of continual learning methods between the i-Blurry scenario and our Si-Blurry scenario on the CIFAR-100 \cite{krizhevsky2009learning} dataset. `MVP-R (2,000)' indicates our method with a memory buffer size of 2,000. The worst and best case denote the lowest and highest score respectively among the 5 independent runs, and the average score is the $A_{Last}$ score.}
\label{table-I_verses_SI}

\centering
\begin{tabular}{ccccccc}
\specialrule{1.1pt}{1pt}{1pt}
\multirow{2}{*}{Disjoint Class Ratio}\rule{0pt}{10pt} & \multicolumn{2}{c}{0} & \multicolumn{2}{c}{50} & \multicolumn{2}{c}{100} \\ \cline{2-7} 
 & $A_\mathrm{AUC}$\rule{0pt}{10pt} & $A_\mathrm{Last}$ & $A_\mathrm{AUC}$ & $A_\mathrm{Last}$ & $A_\mathrm{AUC}$ & $A_\mathrm{Last}$ \\ \hline
DualPrompt\rule{0pt}{10pt} \cite{wang2022dualprompt} & \textbf{68.85±2.77} & 72.31±9.19 & 67.07±4.16 & 56.82±3.49  & 71.45±1.67 & 48.68±3.47 \\
\textbf{MVP (Ours)} & 67.86±2.62 & \textbf{73.83±8.34} & \textbf{68.10±4.91} & \textbf{62.59±2.38}   &  \textbf{73.35±2.63} & \textbf{53.40±5.49}  \\ 
\specialrule{1.1pt}{1pt}{1pt}
\end{tabular}
    \vspace{-2mm}
    \caption{We compared the performance of MVP and DualPrompt at different disjoint class ratios. Experimental results show that MVP can be robust to changing disjoint class ratios.}
    \label{table-disjoint}
\end{table*}

\begin{table*}[h!]
\centering
\begin{tabular}{ccccccc}
\specialrule{1.1pt}{1pt}{1pt}
\multirow{2}{*}{Blurry Sample Ratio}\rule{0pt}{10pt} & \multicolumn{2}{c}{10} & \multicolumn{2}{c}{30} & \multicolumn{2}{c}{50} \\ \cline{2-7} 
 & $A_\mathrm{AUC}$\rule{0pt}{10pt} & $A_\mathrm{Last}$ & $A_\mathrm{AUC}$ & $A_\mathrm{Last}$ & $A_\mathrm{AUC}$ & $A_\mathrm{Last}$ \\ \hline
DualPrompt\rule{0pt}{10pt} \cite{wang2022dualprompt} & 67.07±4.16 & 56.82±3.49 & 70.58±2.05 & 59.47±7.38 & 68.08±5.56 & 49.93±2.82 \\
\textbf{MVP (Ours)} & \textbf{68.10±4.91} & \textbf{62.59±2.38} & \textbf{71.10±2.10} & \textbf{63.02±6.68} & \textbf{70.58±2.05} & \textbf{59.47±7.38} \\ 
\specialrule{1.1pt}{1pt}{1pt}
\end{tabular}
    \vspace{-2mm}
    \caption{We compared the performance of MVP and DualPrompt at different blurry sample ratios. Experimental results show that MVP can be more robust to a number of blurry samples than the existing method.}
    \vspace{-2mm}
    \label{table-blurry}
\end{table*}

\subsection{Ablation Study}
Table \ref{table-ablation} shows an ablation study on CIFAR-100 for the proposed components in our method. We set the ablation study whether using a memory buffer or not. We set the Dual Prompt as a baseline for both cases. The ablation study showed that each component of our method was beneficial to the performance.

As shown in the upper table from Table \ref{table-ablation}, instance-wise logit masking and contrastive visual prompt tuning loss scored a better performance without memory than the performance of the baseline in both $A_\mathrm{AUC}$ and $A_\mathrm{last}$. Moreover, using both of them scored better performance than the performance of each component. It implies instance-wise logit masking and contrastive visual prompt tuning loss makes a complementary performance.
Instance-wise logit masking helps to enhance accuracy by making the task easy by masking irrelevant classes. So, the model can learn representative knowledge more efficiently.

We also investigated the GSF loss and AFS in the lower table from Table \ref{table-ablation}. Using both GSF and ASF scored the highest performance among the performance of the lower table. It showed that Each of GSF and AFS can alleviate class imbalance. In addition, using both of them was more helpful to class imbalance considerably. The lower table implies that class imbalance is a crucial problem for prediction performance in Si-Blurry. 

In Table \ref{table-best_ablation}, We compared the set of components for the inter and intra-task forgettings (instance-wise logit masking and contrastive visual prompt tuning loss) with the set of components for the class imbalance (GSF and AFS). When we used all the components, we observed the performance improvement from 67.07±4.16 to 68.10±4.91 in $A_{AUC}$ and from 56.82±3.49 to 62.59±2.38 in $A_{Last}$. This showed that our methods are synergistic.

\subsection{Comparison between i-Blurry and Si-Blurry}
The comparative analysis, as illustrated in Table \ref{table-I_verses_SI}, encompasses a comprehensive evaluation of the best and worst performance metrics for both existing methods and our proposed approach across two distinct scenarios: i-Blurry \cite{koh2021online} and Si-Blurry. In the context of the i-Blurry scenario, the results underscore the superiority of our proposed MVP method, outperforming other existing methodologies in both the best and worst case scenarios. This demonstrates the consistency and robustness of our method in i-Blurry settings. Notably, our method even surpasses the state-of-the-art CLIB method, establishing its prowess in challenging real-world scenarios.

Furthermore, the insights drawn from our evaluation extend beyond the i-Blurry scenario. Our approach yields improved performance not only in the i-Blurry context but also in the Si-Blurry scenario. It is worth highlighting that while some existing methods exhibited a decline in performance within the Si-Blurry scenario, it is indicative of the increased complexity and realism inherent in Si-Blurry, rendering it a more intricate challenge than the i-Blurry scenario. The robust performance in both i-Blurry and Si-Blurry scenarios underscores the adaptability and effectiveness of our method in tackling varying levels of complexity and realism.

\subsection{Disjoint Sample Ratio}
We tested our method at various disjoint class ratios. We compared the performance of the model in the extreme cases of disjoint only and no disjoint, following experiment of \cite{koh2021online}. Table \ref{table-disjoint} demonstrated that the MVP method maintained a robust and high performance across various disjoint class ratios. This suggested that our method is capable of achieving excellent results in both existing blurry and class incremental learning scenarios.

\subsection{Blurry Sample Ratio}
We tested our method at various blurry sample ratios, following the experiment of \cite{koh2021online}. Table \ref{table-blurry} refers that the MVP method maintained robust high performance while the blurry sample increased to half. Considering these results with the results from Table \ref{table-blurry}, we could conclude that our method has the ability to improve performance in a variety of data distributions that we might encounter in the real world. This showed that the MVP method has the ability to effectively solve real-world problems.

\section{Conclusion}
In this paper, we found that the previous CL scenarios fall short in their efforts to reflect reality. We designed a new scenario, Si-blurry, to simulate the complexity of a real-world data stream. In the Si-Blurry scenario, we could find two main branches of problems that degrade online learning performance: intra-task and inter-task forgetting, and class imbalance. To mitigate these problems in the Si-Blurry scenario, we proposed a Mask and Visual Prompt tuning (MVP) consisting of instance-wise logit masking, contrastive visual prompt tuning loss, gradient similarity-based focal loss, and adaptive feature scaling. Our method in the Si-Blurry scenario outperforms the existing CL methods. We believe that the Si-Blurry scenario is a step-forward scenario that reflects the real-world scenario.

Although our MVP shows a standing-out performance, it has some limitations. Different from prior work \cite{wang2022learning}, we selected only one prompt from the pool to facilitate mask learning and prevent the key converging problem. However, selecting multiple keys can achieve knowledge sharing and reduce the risk of missed selections. It suggests further work to train instance-wise masks taking these advantages. Also, batch-wise calculation of the ignorance score makes the method sensitive to batch size. Batch-agnostic online learning is suggested as another further work. 


\section*{Acknowledgement}
This work was supported by the National Research Foundation of Korea (NRF) grant funded by the Korea government (MSIT) (No. 2021R1G1A1094379), and in part by MSIT (Ministry of Science and ICT), Korea, under the ITRC(Information Technology Research Center) support program (IITP-2023-RS-2023-00258649) supervised by the IITP (Institute for Information \& Communications Technology Planning \& Evaluation), and in part by the Institute of Information and Communications Technology Planning and Evaluation (IITP) grant funded by the Korea Government (MSIT) (Artificial Intelligence Innovation Hub) under Grant 2021-0-02068, and by Institute of Information \& communications Technology Planning \& Evaluation 
(IITP) grant funded by the Korea government (MSIT) (No.RS-2022-00155911, Artificial Intelligence Convergence Innovation Human Resources Development (Kyung Hee University)).

{\small
\bibliographystyle{ieee_fullname}
\bibliography{egbib}
}

\clearpage

\maketitle
\appendix

\section{Details on the Compared Methods}
In our experiments involving memory management, we utilized reservoir sampling as our method for memory management. We followed ER \cite{rolnick2019experience} to utilize memory in training, which combines half of the training batch from the streamed data with half of the training batch from memory. As online continual learning can not handle whole data in a task, other memory management methods such as herding selection \cite{rebuffi2017icarl} and mnemonics \cite{liu2020mnemonics} are inapplicable. Also, memory management of Rainbow Memory \cite{bang2021rainbow} is inapplicable in online continual learning. Because they are based on the information of uncertainty from the whole task samples. Thus, we followed the rainbow memory training process from the CLIB \cite{koh2021online}.

LwF \cite{li2017learning} is a classical method in continual learning which leverages knowledge distillation to prevent the model from catastrophic forgetting. LwF was introduced for offline learning. So, we modified the LwF to apply to online continual learning. Modified LwF distills the knowledge in every batch. 

\section{Additional Ablation Studies}
We conducted ablation studies for the hyperparameter $\gamma$, $\textit{m}$ and $\alpha$ value used in gradient similarity-based focal loss, adaptive feature scaling, and total loss respectively. 

\begin{table}[]
\begin{center}
    
\resizebox{0.8\columnwidth}{!}{
\begin{tabular}{cccc}
\specialrule{1.1pt}{1pt}{1pt}
Method                      & $\gamma$        & $A_\mathrm{AUC}$                 & $A_\mathrm{Last}$                \\ \hline
Baseline\rule{0pt}{10pt}    & - & 67.07±4.16 & 56.82±3.49 \\ \hline
\multirow{5}{*}{MVP (Ours)} & 0.5\rule{0pt}{10pt}& 67.25±5.08          & 60.39±1.55          \\
                            & 1.0          & 67.45±5.05          & 60.95±1.61          \\
                            & 1.5          & 67.52±5.11          & 61.05±1.37          \\
                            & \underline{\textbf{2.0}} & \textbf{68.10±4.91} & \textbf{62.59±2.38} \\
                            & 2.5          & 67.62±5.17          & 61.11±1.55         \\
\specialrule{1.1pt}{1pt}{1pt}
\end{tabular}%
}
\caption{$\gamma$ controls the loss value of gradient similarity-based focal lass. Underlined value denotes the used value for our method and the bold value represents the highest performance in the table.}
\label{table-gamma}
\end{center}

\begin{center}
\resizebox{0.8\columnwidth}{!}{
\begin{tabular}{cccc}
\specialrule{1.1pt}{1pt}{1pt}

Method                      & $\textit{m}$       & $A_\mathrm{AUC}$                 & $A_\mathrm{Last}$                \\ \hline
Baseline\rule{0pt}{10pt}    & - & 67.07±4.16 & 56.82±3.49 \\ \hline
\multirow{5}{*}{MVP (Ours)} & 0.1\rule{0pt}{10pt}& 67.20±4.72          & 58.82±1.27          \\
                            & 0.3          & 67.49±4.83          & 60.04±1.12          \\
                            & \underline{\textbf{0.5}}          &  \textbf{68.10±4.91}         &  \textbf{62.59±2.38}    \\
                            & 0.7          & 67.89±4.94          & 61.31±1.71          \\
                            & 0.9 & 67.29±4.84 & 61.81±0.47 \\
\specialrule{1.1pt}{1pt}{1pt}
\end{tabular}%
}
\caption{$\textit{m}$ is a margin value used in calculating the marginal benefit score by a sample. Underlined value denotes the used value for our method and the bold value represents the highest performance in the table.}
\label{table-margin}
\end{center}
\centering
\resizebox{\columnwidth}{!}{
\begin{tabular}{ccccc}
    \specialrule{1.1pt}{1pt}{1pt}
                 \multirow{2}{*}{$\alpha$} & \multicolumn{2}{c}{\textbf{Memory = 0}}            & \multicolumn{2}{c}{\textbf{Memory = 2,000}}   \\ \cline{2-3} \cline{4-5}
         & $A_\mathrm{AUC}$    & $A_\mathrm{Last}$   & $A_\mathrm{AUC}$ & $A_\mathrm{Last}$ \\ \hline
    -          & 40.11±1.27          & 29.24±4.63          & 49.00±2.06       &  37.96±0.34      \\ \cdashline{1-5}
    0.1          & 40.38±1.67          & 31.63±3.39          & 52.13±0.14       & 50.50±3.11        \\
    0.3          & 40.52±1.59          & 31.81±3.66          & 52.14±0.28       & 50.51±2.76        \\
    \underline{\textbf{0.5}} & \textbf{40.60±1.21} & \textbf{31.96±3.07} & \textbf{52.47±1.45}       & \textbf{50.54±2.08}        \\
    0.7          & 40.53±1.01          & 31.56±2.05          & 52.28±2.34       & 50.43±1.53        \\
    \specialrule{1.1pt}{1pt}{1pt}
\end{tabular}%
}
\captionof{table}{ $\alpha$ is a balancing value in total losses. Underlined value denotes the used value for our method and the bold value represents the highest performance in the table.}
\label{table-alpha}

\end{table}

\subsection{Hyperparameters $\gamma$ and $\textit{m}$}
Table \ref{table-gamma} shows the result of the hyperparameter $\gamma$ ablation study. Hyperparameter $\gamma$ controls the ignore score calculated by a sample in gradient similarity-based focal loss. We set the $\gamma$ value to 0.5, 1.0, 1.5, 2.0, 2.5 and evaluate the performance by $A_\mathrm{AUC}$ and $A_\mathrm{Last}$. When the $\gamma$ was 2.0, our method scored optimal performance in both $A_\mathrm{AUC}$ and $A_\mathrm{Last}$. Also, we experimented with the performance variation by the hyperparameter $\textit{m}$. Hyperparameter $\textit{m}$ is used in adaptive feature scaling to yield a marginal benefit score of a sample. Table \ref{table-margin} shows the result of the performance variation by the hyperparameter $\textit{m}$. As we could see through Table \ref{table-margin}, our novel method was robust to margin value. So, we set the $\textit{m}$ to 0.5 showing the highest performance among all seeds.  

\subsection{Hyperparameter $\alpha$}
Table \ref{table-alpha} presents the performance of our method for various values of $\alpha$. Notably, the optimal performance is observed when an $\alpha$ value was 0.5, demonstrating consistent and robust results across different $\alpha$ values. Based on these findings, we set the $\alpha$ value to 0.5 as the most suitable choice for our experiments. This decision is grounded in the stability and high performance exhibited by the method at this particular alpha value, ensuring reliable and reproducible outcomes in our research.


\subsection{Mask-Prompt Pool Size and Prompt Selection}

\begin{figure}[]
\begin{center}
    \centering
    \includegraphics[width=\columnwidth]{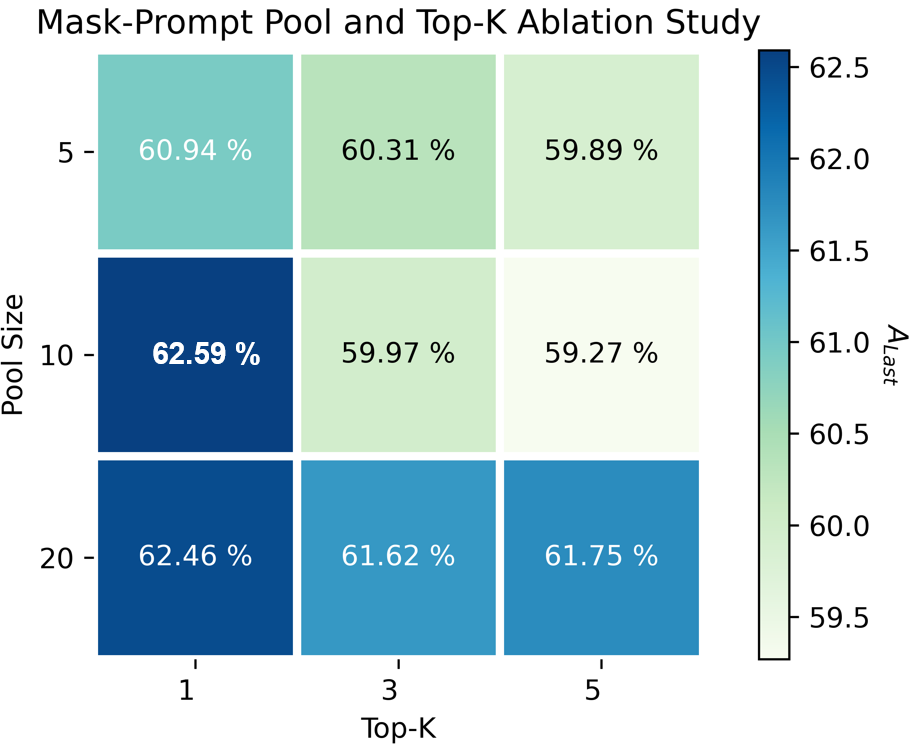}
\vspace{-0.2cm}
\caption{We set the prompt pool size and the number of selected prompts to 5, 10, and 20 and 1, 3, and 5 respectively. Top-K denotes the number of selected prompts.}
\label{fig:prompt_selection}
\end{center}
\end{figure}

\begin{table}
\centering
\begin{tabular}{ll}
\specialrule{1.1pt}{1pt}{1pt}
   \textbf{Method}  & \textbf{Forgetting} \\
\hline
    Baseline & 46.61±5.30 \\ \cdashline{1-2}
    \ \ + GSF,AFS     & 45.35±4.40 \\
    \ \ + Cont,Mask     & 39.98±4.02 \\
    \ \ + Cont,Mask,GSF,AFS & \textbf{39.68±3.98} \\
\specialrule{1.1pt}{1pt}{1pt}
\end{tabular}
\captionof{table}{The ablation study on forgetting on ImageNet-R. The results demonstrate that our approach significantly mitigates forgetting and ensures better retention of previously learned knowledge.}
\label{tab:forgetting}
\end{table}

Since the number of mask-prompt pairs has a large impact on our method, we ran experiments with a variety of mask-prompt pool sizes and prompt selection. Figure \ref{fig:prompt_selection} represents the $A_\mathrm{Last}$ scores from the mask-prompt pool size and the number of selections. Top-K denotes the number of selected prompts. As shown in this figure, when the prompt pool size was fixed, a performance drop happened when more prompts were selected. Since selecting more masks and prompts induced much severe forgetting in each prompt, selecting a lot of masks and prompts exacerbated the performance. We set the mask-prompt pool size to 10 and the number of selection sizes to 1 to ensure the optimal performance of our method.

\subsection{Forgetting}
As shown in Table~\ref{tab:forgetting}, we conducted experiments to assess the impact of each method on forgetting. Our findings revealed that GSF and AFS had limited effects on forgetting, as they predominantly targeted minor and major classes, respectively, in the class imbalance scenario. In contrast, our proposed approach, contrastive prompt tuning, demonstrated significant effectiveness in addressing the challenges of key floating and selection. Additionally, the utilization of masking proved to be highly effective in preventing forgetting by inhibiting the backpropagation of fully learned knowledge. These results collectively emphasize the robustness and efficiency of our method in effectively mitigating forgetting during the learning process.

\section{Visualization of Masks and Prompt Keys}
We performed visualizations to verify the suggested method experimentally and to understand our novel method MVP further. We visualized the mask and key of prompt methods.

\begin{figure}[]
\begin{center}
    \subfigure[Visualization of each mask value from class 0 to 9 after training task 0]{
        \centering
        \includegraphics[width=\columnwidth]{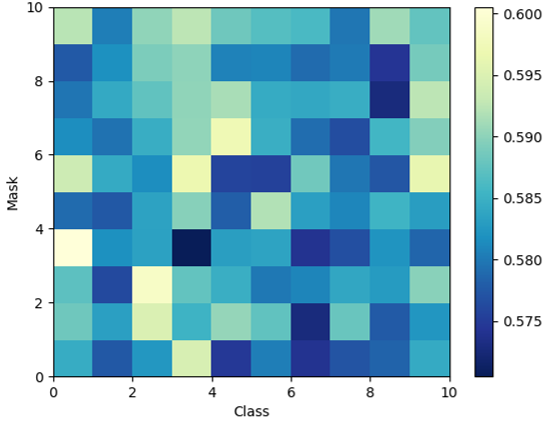}
    }
    \subfigure[Visualization of each mask value from class 0 to 9 after training task 4]{
        \centering
        \includegraphics[width=\columnwidth]{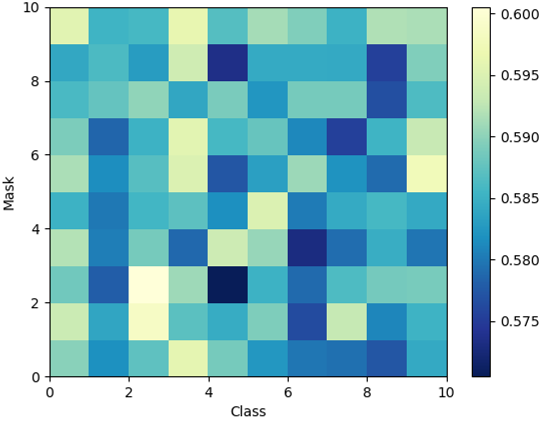}
    }
\end{center}
    \caption{Visualization of mask value from class 0 to 9 (a) after task 0 (b) after task 4. Because each mask blocks logit from a different class, it seems to be noisy. It could be observed the value of the mask change as it trained.}
    \label{fig:mask}
\end{figure}

\subsection{Instance-wise Logit Mask}

In order to validate the effectiveness of the mask used in our proposed method, we conducted a mask visualization experiment. The purpose of this experiment was to gain a better understanding of how the mask is utilized during the learning process. As Figure \ref{fig:mask} illustrates, we could see that each mask opens for a different class. We could also see that on some parts of the masks, classes had their values decreased, preventing further updates. Also, some of them got increased value, allowing the model to learn relevant knowledge. The results demonstrated that the mask is effective in facilitating the division of tasks and protecting knowledge as intended. By facilitating the division of tasks and protecting knowledge, the mask enabled our method to perform well even in scenarios with blurry boundaries and multiple classes in a single batch. This is a significant contribution to the field of continuous learning and has important implications for real-world applications.

\subsection{Prompt Key}

\begin{figure}[]
\begin{center}
    \subfigure[t-SNE Visualization of prompt key of DualPrompt \cite{wang2022dualprompt} after training task 0]{
        \centering
        \includegraphics[width=0.47\columnwidth]{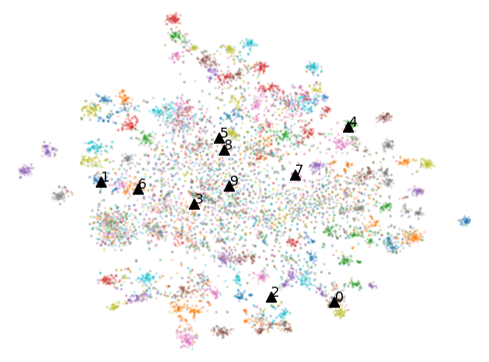}
    }
    \subfigure[t-SNE Visualization of prompt key of DualPrompt \cite{wang2022dualprompt} after training task 4]{
        \centering
        \includegraphics[width=0.47\columnwidth]{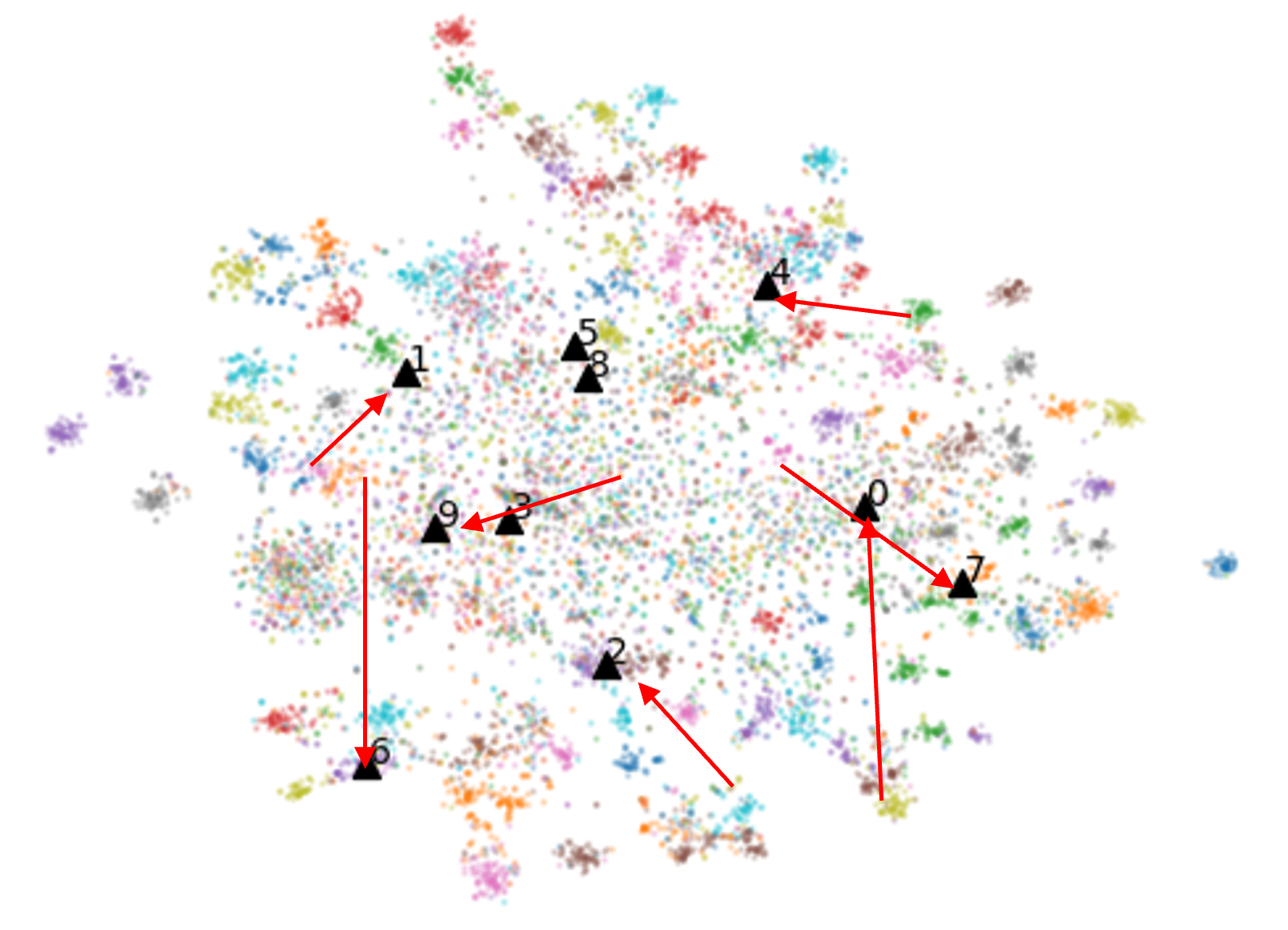}
    }
    \subfigure[t-SNE Visualization of prompt key of MVP after training task 0]{
        \centering
        \includegraphics[width=0.47\columnwidth]{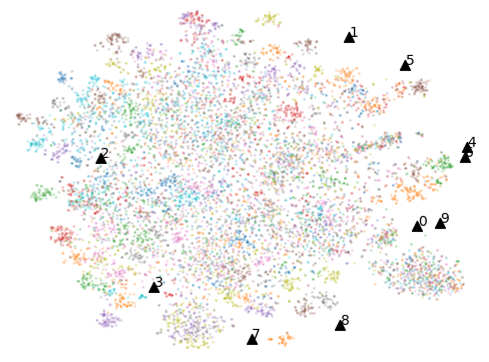}
    }
    \subfigure[t-SNE Visualization of prompt key of MVP after training task 4]{
        \centering
        \includegraphics[width=0.47\columnwidth]{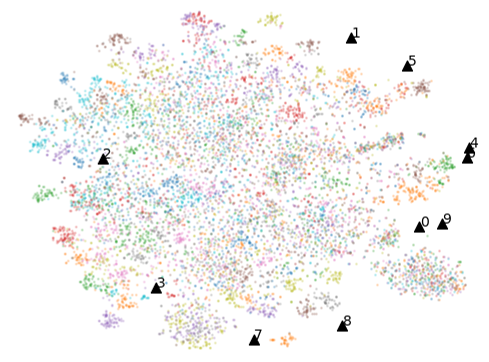}
    }
\end{center}
\vspace{-0.5cm}
    \caption{t-SNE visualization of the prompt key of \iffalse(a) L2P \cite{wang2022learning} after task 0 (b) L2P \cite{wang2022learning} after task 4 \fi(a) DualPrompt \cite{wang2022dualprompt} after task 0 (b) Dualprompt \cite{wang2022dualprompt} after task 4 (c) MVP after task 0 (d) MVP after task 4. \iffalse L2P exhibits the phenomenon of keys converging to a single point. \fi DualPrompt suffered from the semantic drift because the key is constantly changing as the task changes.}
    \label{fig:tSNE}
\end{figure}

Figure \ref{fig:tSNE} shows the t-SNE visualization of the prompt key used in each prompt-based methods. Since DualPrompt \cite{wang2022dualprompt} do not have any constraints on key learning in common, we saw the keys floating as the task changes. This changes the function of each prompt in the feature space and could cause severe semantic drift. In MVP, we could see that the keys are kept at a reasonable distance from each other and the movement is suppressed once learning is sufficiently advanced.

\begin{table}[]
\begin{center}
\resizebox{\columnwidth}{!}{%
\begin{tabular}{cccc}
\specialrule{1.1pt}{1pt}{1pt}
\multirow{2}{*}{Memory Size}\rule{0pt}{10pt} & \multirow{2}{*}{Methods} & \multicolumn{2}{c}{Metrics}               \\ \cline{3-4} 
                             &                          & $A_\mathrm{Last}$ (↑)\rule{0pt}{10pt}& $\mathrm{Forgetting}$ (↓)      \\ \hline
\multirow{5}{*}{0}           & FineTuning\rule{0pt}{10pt} & 10.42±4.92          & 45.11±5.98          \\
                             & LwF \cite{li2017learning}   & 36.53±10.96         & 56.43±12.91         \\
                             & L2P \cite{wang2022learning}              & 41.63±12.73         & 55.46±13.15         \\
                             & DualPrompt \cite{wang2022dualprompt}  & 56.82±3.49          & 40.35±1.25          \\
                             & \textbf{MVP (Ours)}            & \textbf{62.59±2.38} & \textbf{34.63±2.46} \\ \hline
\multirow{5}{*}{500}         & ER\rule{0pt}{10pt} \cite{rolnick2019experience} & 60.68±1.15          & 28.85±3.51          \\
                             & EWC++ \cite{kirkpatrick2017overcoming} & 25.62±3.35          & 47.16±9.72          \\
                             & RM \cite{bang2021rainbow} & 23.94±0.61          & 24.28±2.90          \\
                             & CLIB \cite{koh2021online} & 67.16±0.72          & 15.45±0.94          \\
                             & \textbf{MVP (Ours)}            & \textbf{79.32±1.28} & \textbf{14.57±1.60} \\ \hline
\multirow{5}{*}{2000}        & ER\rule{0pt}{10pt} \cite{rolnick2019experience}& 71.81±0.69          & 15.45±0.94          \\
                             & EWC++ \cite{kirkpatrick2017overcoming} & 46.93±1.44          & 28.75±7.58          \\
                             & RM \cite{bang2021rainbow} & 65.51±0.55          & 9.50±1.49         \\
                             & CLIB \cite{koh2021online} & 72.09±0.49          & \textbf{8.07±0.98}          \\
                             & \textbf{MVP (Ours)}            & \textbf{84.42±0.44} & 8.79±1.49 \\
\specialrule{1.1pt}{1pt}{1pt}
\end{tabular}%
}
\caption{We compared our method, MVP, to other existing methods in two metrics. Forgetting is measured with the best accuracy of each class and the inference accuracy after all the tasks are trained.}
\label{table-forget}
\end{center}
\vspace{-0.5cm}
\end{table}

\section{Discussions}

\begin{table*}[h]
\begin{minipage}{\textwidth}
\parbox{0.65\textwidth}{
\begin{tabular}{ccccc}
    \specialrule{1.1pt}{1pt}{1pt}
    \multirow{2}{*}{\textbf{Method}} & \multicolumn{2}{c}{\textbf{Memory = 500}} & \multicolumn{2}{c}{\textbf{Memory = 2,000}} \\ \cline{2-3} \cline{4-5}
                            & $A_\mathrm{AUC}$    & $A_\mathrm{Last}$   & $A_\mathrm{AUC}$ & $A_\mathrm{Last}$ \\ \hline
    L2P                     & 69.91±1.49              & 56.58±0.64              & 75.24±0.82       & 68.73±0.80      \\
    DualPrompt              & 75.07±1.01               & 62.12±1.50              & 79.76±0.47       & 72.09±0.80      \\    \cdashline{1-5}
    \textbf{MVP-R (Ours)}   & \textbf{76.52±0.73}      & \textbf{65.19±0.58}     & \textbf{80.67±0.75}       & \textbf{74.34±0.32}     \\
    \specialrule{1.1pt}{1pt}{1pt}
\end{tabular}
\vspace{-2mm}
\captionof{table}{Comparison of ours with L2P and DualPrompt on Tiny ImageNet.}
\label{tab:memory}
\vspace{1.5mm}
}
\parbox{0.35\textwidth}{
\begin{tabular}{ccc}
    \specialrule{1.1pt}{1pt}{1pt}
        \textbf{Method} & \textbf{TFLOPs} & \makecell{\bf Training (s)\\\bf /Iter} \\
        \hline
        CLIB & 69.6 & 11.590 \\
        DualPrompt & 4.37 & 0.906 \\ \cdashline{1-3}
        \textbf{MVP (Ours)} & \textbf{4.19} & \textbf{0.882} \\
    \specialrule{1.1pt}{1pt}{1pt}
\end{tabular}
\vspace{-2mm}
\captionof{table}{Computational cost Analysis of each method.}
\label{tab:copmute}
\vspace{-3mm}
}
\end{minipage}
\end{table*}

\subsection{Additional Results for the Forgetting Score}
Table \ref{table-forget} shows the performance of our proposed method with respect to the accuracy score and forgetting score. We used the forgetting measurement in \cite{chaudhry2018riemannian} to report the forgetting results. As shown in this table, our method not only scored the highest accuracy in the table but also the lowest forgetting score. It denoted that MVP performs at best in  accuracy while minimizing the forgetting knowledge.

Note that low forgetting score do not mean a better method than others. If a model did not train with newly streamed data, there is no forgetting. However, reporting the low forgetting score while keeping the high prediction accuracy represents that the model can capture the knowledge from the new data while preventing the model from forgetting existing knowledge. Thus, forgetting measurement considering prediction accuracy is crucial to estimate the stability-plasticity of the method.

\subsection{Additional Results with Memory}
L2P and DualPrompt were initially not explicitly designed to incorporate memory, although they can be utilized in conjunction with memory. As shown in Table~\ref{tab:memory}, we evaluated their performance in the presence of additional memory. Through extensive experiments conducted on the Tiny-ImageNet Dataset, we observed that our method significantly surpassed DualPrompt and L2P. This compelling outcome confirms that the performance enhancement achieved by our method over the baseline is attributed to additional factors brought into play by memory utilization. These findings reinforce the effectiveness and advantages of our approach, particularly when memory is incorporated, leading to notable improvements in performance compared to the baseline methods.

\begin{figure}[t]
\begin{center}
            \includegraphics[width=\columnwidth]{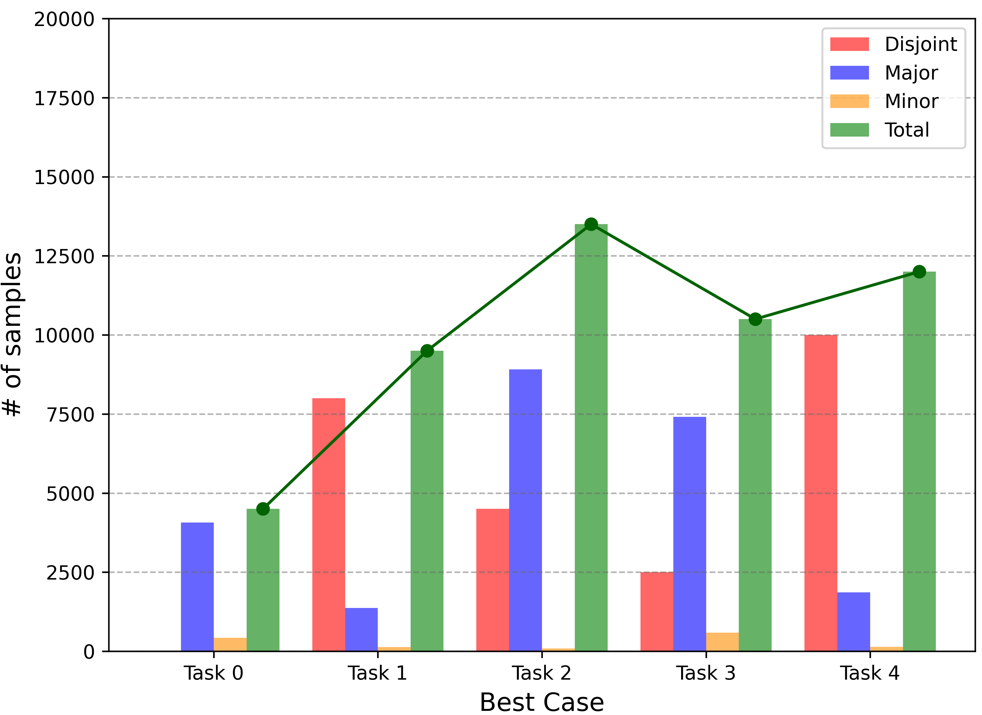}
\end{center}
    \caption{This figure represents the task configuration of training data in the best case. We reported the number of samples from each task. Total means the summation of training samples. We observed that training data are impartially distributed among the tasks in the best case.}
    \label{fig:best}

\begin{center}
    \includegraphics[width=\columnwidth]{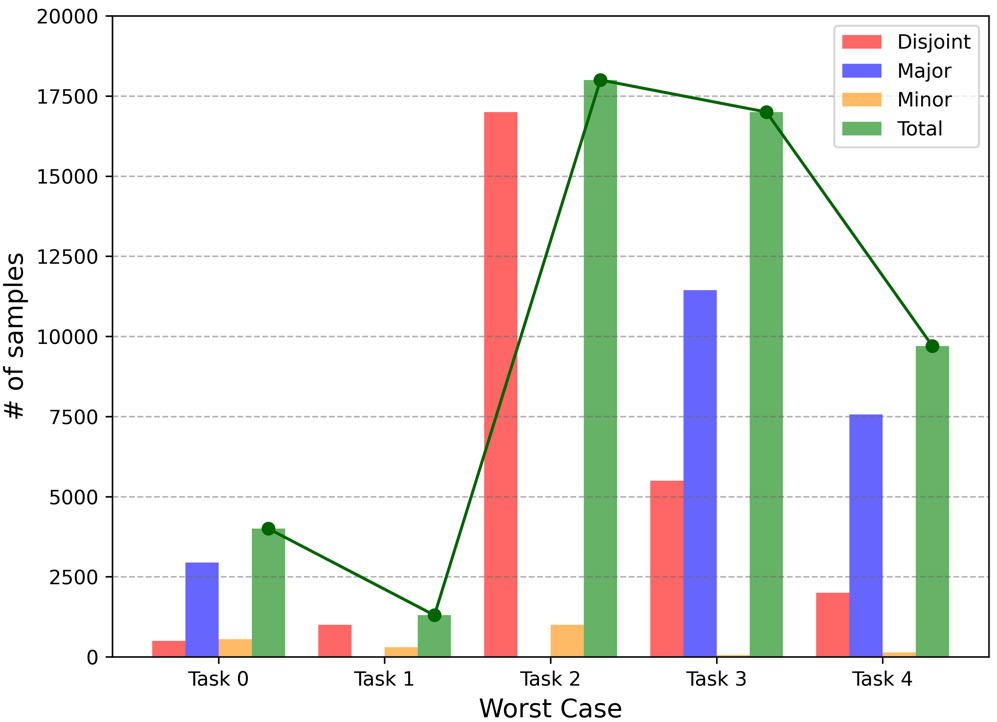}
\end{center}
    \caption{The above figure represents the task configuration of the worst case. Total means the summation of training samples. We observed that training data were concentrated on some tasks in the worst case.}
    \label{fig:worst}
\end{figure}

\subsection{Computational Cost}
In Table~\ref{tab:copmute}, we conducted a thorough analysis of the computational cost associated with each method. This analysis encompassed a comparison of all methods using a memory capacity of 2000. Notably, the CLIB method necessitates forwarding for every individual sample to calculate the memory importance, resulting in a substantial computational overhead. In contrast, our method achieves a lower computational cost in comparison to DualPrompt (DP) by strategically reducing certain operations during the prompt selection process.


\begin{figure}
    \centering
    \includegraphics[width=0.8\columnwidth]{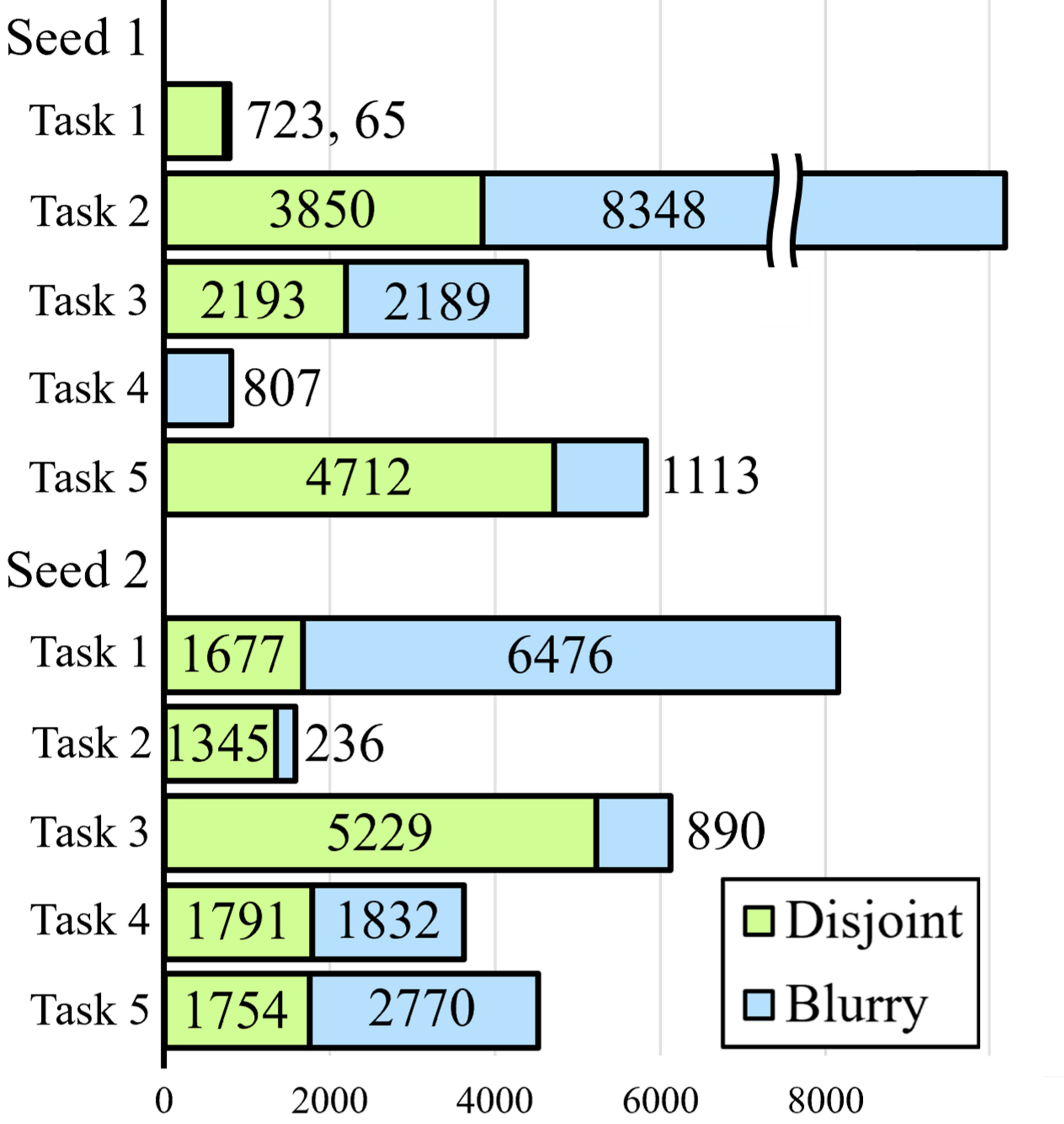}
    \caption{Example of Si-Blurry Scenario.}
    \label{fig:siblurry}
\end{figure}

\subsection{Task Configuration of Best and Worst Cases}

We classified the classes into 3 categories: disjoint, major, and minor in the Si-Blurry scenario. Disjoint classes mean newly incomed classes that never appeared before. Since disjoint classes appear only once with all the training data, there is no overlap between tasks. Major classes and minor classes have a blurry task boundary. If a major class appeared in a task, that class turn to minor classes in other tasks. Hence, the major class can be overlapped between the tasks and once the major class appeared, it becomes the minor class in other tasks.
 
The Figure \ref{fig:siblurry} shows another variety of possible Si-Blurry scenarios. Among these possibilities, we analysed the highest and lowest performing cases. Figure \ref{fig:best} and Figure \ref{fig:worst} show the task configuration when our method scored the highest and lowest performance among the all seeds. It is noteworthy that the task configuration in the best case seemed like training samples are distributed impartially and in the worst case, training samples are concentrated on some tasks.

In other words, in the best case, the training data were impartially distributed to all tasks and it leaded to relatively low biased in tasks. The model could learn the knowledge among the tasks without severe weight drift or biased to some tasks. In the worst case, however, the training data were highly focused on some tasks and it leaded to a different amount of learning in the model training between tasks. In this case, the model could suffer severe catastrophic forgetting \cite{MCCLOSKEY1989109, FRENCH1999128} and be highly biased to the tasks with a lot of training samples. Our novel method MVP resolved this bias problem, showed better result than prior works. 

{\small
\bibliographystyle{ieee_fullname}
\bibliography{egbib}

\begin{thebibliography}{10}\itemsep=-1pt

\bibitem{aljundi2017expert}
Rahaf Aljundi, Punarjay Chakravarty, and Tinne Tuytelaars.
\newblock Expert gate: Lifelong learning with a network of experts.
\newblock In {\em CVPR}, 2017.

\bibitem{NEURIPS2019_e562cd9c}
Rahaf Aljundi, Min Lin, Baptiste Goujaud, and Yoshua Bengio.
\newblock Gradient based sample selection for online continual learning.
\newblock In H. Wallach, H. Larochelle, A. Beygelzimer, F. d\textquotesingle
  Alch\'{e}-Buc, E. Fox, and R. Garnett, editors, {\em Advances in Neural
  Information Processing Systems}, volume~32. Curran Associates, Inc., 2019.

\bibitem{bang2021rainbow}
Jihwan Bang, Heesu Kim, YoungJoon Yoo, Jung-Woo Ha, and Jonghyun Choi.
\newblock Rainbow memory: Continual learning with a memory of diverse samples.
\newblock In {\em CVPR}, 2021.

\bibitem{castro2018end}
Francisco~M Castro, Manuel~J Mar{\'\i}n-Jim{\'e}nez, Nicol{\'a}s Guil, Cordelia
  Schmid, and Karteek Alahari.
\newblock End-to-end incremental learning.
\newblock In {\em ECCV}, 2018.

\bibitem{chaudhry2018riemannian}
Arslan Chaudhry, Puneet~K Dokania, Thalaiyasingam Ajanthan, and Philip~HS Torr.
\newblock Riemannian walk for incremental learning: Understanding forgetting
  and intransigence.
\newblock In {\em ECCV}, 2018.

\bibitem{chaudhry2019tiny}
Arslan Chaudhry, Marcus Rohrbach, Mohamed Elhoseiny, Thalaiyasingam Ajanthan,
  Puneet~K Dokania, Philip~HS Torr, and Marc'Aurelio Ranzato.
\newblock On tiny episodic memories in continual learning.
\newblock {\em arXiv preprint arXiv:1902.10486}, 2019.

\bibitem{cui2019class}
Yin Cui, Menglin Jia, Tsung-Yi Lin, Yang Song, and Serge Belongie.
\newblock Class-balanced loss based on effective number of samples.
\newblock In {\em CVPR}, 2019.

\bibitem{de2021continual}
Matthias De~Lange, Rahaf Aljundi, Marc Masana, Sarah Parisot, Xu Jia,
  Ale{\v{s}} Leonardis, Gregory Slabaugh, and Tinne Tuytelaars.
\newblock A continual learning survey: Defying forgetting in classification
  tasks.
\newblock {\em TPAMI}, 2021.

\bibitem{deng2019arcface}
Jiankang Deng, Jia Guo, Niannan Xue, and Stefanos Zafeiriou.
\newblock Arcface: Additive angular margin loss for deep face recognition.
\newblock In {\em CVPR}, 2019.

\bibitem{dosovitskiy2020image}
Alexey Dosovitskiy, Lucas Beyer, Alexander Kolesnikov, Dirk Weissenborn,
  Xiaohua Zhai, Thomas Unterthiner, Mostafa Dehghani, Matthias Minderer, Georg
  Heigold, Sylvain Gelly, et~al.
\newblock An image is worth 16x16 words: Transformers for image recognition at
  scale.
\newblock {\em arXiv preprint arXiv:2010.11929}, 2020.

\bibitem{FRENCH1999128}
Robert~M. French.
\newblock Catastrophic forgetting in connectionist networks.
\newblock {\em Trends in Cognitive Sciences}, 3(4):128--135, 1999.

\bibitem{gepperth2016incremental}
Alexander Gepperth and Barbara Hammer.
\newblock Incremental learning algorithms and applications.
\newblock In {\em ESANN}, 2016.

\bibitem{gupta2020neural}
Pankaj Gupta, Yatin Chaudhary, Thomas Runkler, and Hinrich Schuetze.
\newblock Neural topic modeling with continual lifelong learning.
\newblock In {\em ICML}, 2020.

\bibitem{hendrycks2021many}
Dan Hendrycks, Steven Basart, Norman Mu, Saurav Kadavath, Frank Wang, Evan
  Dorundo, Rahul Desai, Tyler Zhu, Samyak Parajuli, Mike Guo, et~al.
\newblock The many faces of robustness: A critical analysis of
  out-of-distribution generalization.
\newblock In {\em ICCV}, 2021.

\bibitem{jia2022visual}
Menglin Jia, Luming Tang, Bor-Chun Chen, Claire Cardie, Serge Belongie, Bharath
  Hariharan, and Ser-Nam Lim.
\newblock Visual prompt tuning.
\newblock In {\em ECCV}, 2022.

\bibitem{adam}
Diederik~P Kingma and Jimmy Ba.
\newblock Adam: A method for stochastic optimization.
\newblock {\em arXiv preprint arXiv:1412.6980}, 2014.

\bibitem{kirkpatrick2017overcoming}
James Kirkpatrick, Razvan Pascanu, Neil Rabinowitz, Joel Veness, Guillaume
  Desjardins, Andrei~A Rusu, Kieran Milan, John Quan, Tiago Ramalho, Agnieszka
  Grabska-Barwinska, et~al.
\newblock Overcoming catastrophic forgetting in neural networks.
\newblock {\em Proceedings of the national academy of sciences},
  114(13):3521--3526, 2017.

\bibitem{koh2021online}
Hyunseo Koh, Dahyun Kim, Jung-Woo Ha, and Jonghyun Choi.
\newblock Online continual learning on class incremental blurry task
  configuration with anytime inference.
\newblock In {\em International Conference on Learning Representations}.

\bibitem{krizhevsky2009learning}
Alex Krizhevsky.
\newblock Learning multiple layers of features from tiny images.
\newblock pages 32--33, 2009.

\bibitem{le2015tiny}
Ya Le and Xuan Yang.
\newblock Tiny imagenet visual recognition challenge.

\bibitem{lesort2020continual}
Timoth{\'e}e Lesort, Vincenzo Lomonaco, Andrei Stoian, Davide Maltoni, David
  Filliat, and Natalia D{\'\i}az-Rodr{\'\i}guez.
\newblock Continual learning for robotics: Definition, framework, learning
  strategies, opportunities and challenges.
\newblock {\em IF}, 2020.

\bibitem{lester2021power}
Brian Lester, Rami Al-Rfou, and Noah Constant.
\newblock The power of scale for parameter-efficient prompt tuning.
\newblock {\em arXiv preprint arXiv:2104.08691}, 2021.

\bibitem{li2017learning}
Zhizhong Li and Derek Hoiem.
\newblock Learning without forgetting.
\newblock {\em TPAMI}, 2017.

\bibitem{lin2017focal}
Tsung-Yi Lin, Priya Goyal, Ross Girshick, Kaiming He, and Piotr Doll{\'a}r.
\newblock Focal loss for dense object detection.
\newblock In {\em ICCV}, 2017.

\bibitem{liu2017sphereface}
Weiyang Liu, Yandong Wen, Zhiding Yu, Ming Li, Bhiksha Raj, and Le Song.
\newblock Sphereface: Deep hypersphere embedding for face recognition.
\newblock In {\em CVPR}, 2017.

\bibitem{liu2016large}
Weiyang Liu, Yandong Wen, Zhiding Yu, and Meng Yang.
\newblock Large-margin softmax loss for convolutional neural networks.
\newblock {\em arXiv preprint arXiv:1612.02295}, 2016.

\bibitem{liu2020more}
Yu Liu, Sarah Parisot, Gregory Slabaugh, Xu Jia, Ales Leonardis, and Tinne
  Tuytelaars.
\newblock More classifiers, less forgetting: A generic multi-classifier
  paradigm for incremental learning.
\newblock In {\em ECCV}, 2020.

\bibitem{liu2020mnemonics}
Yaoyao Liu, Yuting Su, An-An Liu, Bernt Schiele, and Qianru Sun.
\newblock Mnemonics training: Multi-class incremental learning without
  forgetting.
\newblock In {\em CVPR}, 2020.

\bibitem{lopez2017gradient}
David Lopez-Paz and Marc'Aurelio Ranzato.
\newblock Gradient episodic memory for continual learning.
\newblock volume~30, 2017.

\bibitem{MCCLOSKEY1989109}
Michael McCloskey and Neal~J. Cohen.
\newblock Catastrophic interference in connectionist networks: The sequential
  learning problem.
\newblock volume~24 of {\em Psychology of Learning and Motivation}, pages
  109--165. Academic Press, 1989.

\bibitem{parisi2019continual}
German~I Parisi, Ronald Kemker, Jose~L Part, Christopher Kanan, and Stefan
  Wermter.
\newblock Continual lifelong learning with neural networks: A review.
\newblock {\em NN}, 2019.

\bibitem{pfulb2018a}
B. Pfülb and A. Gepperth.
\newblock A comprehensive, application-oriented study of catastrophic
  forgetting in {DNN}s.
\newblock In {\em ICLR}, 2019.

\bibitem{prabhu2020gdumb}
Ameya Prabhu, Philip~HS Torr, and Puneet~K Dokania.
\newblock Gdumb: A simple approach that questions our progress in continual
  learning.
\newblock In {\em ECCV}, 2020.

\bibitem{rebuffi2017icarl}
Sylvestre-Alvise Rebuffi, Alexander Kolesnikov, Georg Sperl, and Christoph~H
  Lampert.
\newblock icarl: Incremental classifier and representation learning.
\newblock In {\em CVPR}, 2017.

\bibitem{rolnick2019experience}
David Rolnick, Arun Ahuja, Jonathan Schwarz, Timothy Lillicrap, and Gregory
  Wayne.
\newblock Experience replay for continual learning.
\newblock 2019.

\bibitem{rusu2016progressive}
Andrei~A Rusu, Neil~C Rabinowitz, Guillaume Desjardins, Hubert Soyer, James
  Kirkpatrick, Koray Kavukcuoglu, Razvan Pascanu, and Raia Hadsell.
\newblock Progressive neural networks.
\newblock {\em arXiv preprint arXiv:1606.04671}, 2016.

\bibitem{shin2017continual}
Hanul Shin, Jung~Kwon Lee, Jaehong Kim, and Jiwon Kim.
\newblock Continual learning with deep generative replay.
\newblock 2017.

\bibitem{wang2022foster}
Fu-Yun Wang, Da-Wei Zhou, Han-Jia Ye, and De-Chuan Zhan.
\newblock Foster: Feature boosting and compression for class-incremental
  learning.
\newblock In {\em ECCV}, 2022.

\bibitem{wang2022dualprompt}
Zifeng Wang, Zizhao Zhang, Sayna Ebrahimi, Ruoxi Sun, Han Zhang, Chen-Yu Lee,
  Xiaoqi Ren, Guolong Su, Vincent Perot, Jennifer Dy, et~al.
\newblock Dualprompt: Complementary prompting for rehearsal-free continual
  learning.
\newblock In {\em ECCV}, 2022.

\bibitem{wang2022learning}
Zifeng Wang, Zizhao Zhang, Chen-Yu Lee, Han Zhang, Ruoxi Sun, Xiaoqi Ren,
  Guolong Su, Vincent Perot, Jennifer Dy, and Tomas Pfister.
\newblock Learning to prompt for continual learning.
\newblock In {\em CVPR}, 2022.

\bibitem{welling2009herding}
Max Welling.
\newblock Herding dynamical weights to learn.
\newblock In {\em ICMR}, 2009.

\bibitem{wu2019large}
Yue Wu, Yinpeng Chen, Lijuan Wang, Yuancheng Ye, Zicheng Liu, Yandong Guo, and
  Yun Fu.
\newblock Large scale incremental learning.
\newblock In {\em CVPR}, 2019.

\bibitem{yan2021dynamically}
Shipeng Yan, Jiangwei Xie, and Xuming He.
\newblock Der: Dynamically expandable representation for class incremental
  learning.
\newblock In {\em CVPR}, 2021.

\bibitem{yoon2017lifelong}
Jaehong Yoon, Eunho Yang, Jeongtae Lee, and Sung~Ju Hwang.
\newblock Lifelong learning with dynamically expandable networks.
\newblock {\em arXiv preprint arXiv:1708.01547}, 2017.

\end{thebibliography}
}

\end{document}